%% file: PaperForReview.tex
\documentclass[10pt,twocolumn,letterpaper]{article}

\usepackage[pagenumbers]{cvpr} 

\usepackage{graphicx}
\usepackage{amsmath}
\usepackage{amssymb}
\usepackage{booktabs}
\usepackage{cite}
\usepackage{array}
\usepackage{pifont}
\usepackage{float}
\usepackage[pagebackref,breaklinks,colorlinks]{hyperref}
\usepackage{multirow}
\usepackage[table]{xcolor}

\definecolor{lightpurple}{RGB}{200, 230, 245}
\definecolor{lightblue}{RGB}{230, 245, 255}

\usepackage[capitalize]{cleveref}
\crefname{section}{Sec.}{Secs.}
\Crefname{section}{Section}{Sections}
\Crefname{table}{Table}{Tables}
\crefname{table}{Tab.}{Tabs.}

\def\cvprPaperID{235} 
\def\confName{CVPR}
\def\confYear{2025}

\begin{document}

\title{InfoSculpt: Sculpting the Latent Space for Generalized Category Discovery}

\author{
    Wenwen Liao\textsuperscript{1} \quad
    Hang Ruan\textsuperscript{1} \quad
    Jianbo Yu\textsuperscript{2}\thanks{Corresponding author: {\tt\small jb\_yu@fudan.edu.cn}} \quad
    Yuansong Wang\textsuperscript{3} \quad
    Qingchao Jiang\textsuperscript{4} \quad
    Xiaofeng Yang\textsuperscript{2} \\[2mm]
    \textsuperscript{1}College of Intelligent Robotics and Advance Manufacturing, Fudan University\\
    \textsuperscript{2}School of Microelectronics, Fudan University\\
    \textsuperscript{3}Tsinghua Shenzhen International Graduate School, Tsinghua University\\
    \textsuperscript{4}School of Information Science and Engineering, East China University of Science and Technology\\
    {\tt\small wwliao24@m.fudan.edu.cn, jb\_yu@fudan.edu.cn} 
}

\maketitle

\begin{abstract}
Generalized Category Discovery (GCD) aims to classify instances from both known and novel categories within a large-scale unlabeled dataset, a critical yet challenging task for real-world, open-world applications. However, existing methods often rely on pseudo-labeling, or two-stage clustering, which lack a principled mechanism to explicitly disentangle essential, category-defining signals from instance-specific noise. In this paper, we address this fundamental limitation by re-framing GCD from an information-theoretic perspective, grounded in the Information Bottleneck (IB) principle. We introduce \textbf{InfoSculpt}, a novel framework that systematically sculpts the representation space by minimizing a dual Conditional Mutual Information (CMI) objective. InfoSculpt uniquely combines a \textbf{Category-Level CMI} on labeled data to learn compact and discriminative representations for known classes, and a complementary \textbf{Instance-Level CMI} on all data to distill invariant features by compressing augmentation-induced noise. These two objectives work synergistically at different scales to produce a disentangled and robust latent space where categorical information is preserved while noisy, instance-specific details are discarded. Extensive experiments on 8 benchmarks demonstrate that InfoSculpt validating the effectiveness of our information-theoretic approach. Code will be released after acceptance.
\end{abstract}

\section{Introduction}
Deep learning has achieved great success in image classification under the “closed-world” assumption, where all test data belong to known training classes. However, in real-life open-world scenarios \cite{zhu2024open,geng2020recent}, this assumption often fails due to the high cost of data annotation and the dynamic emergence of novel categories. To address this, Novel Category Discovery (NCD) \cite{han2021autonovel} is proposed, but traditional NCD methods assume all unlabeled data come solely from novel classes, which is unrealistic and limits their applicability in practical, complex environments.

\begin{figure}
\centering
\includegraphics[width=0.99\columnwidth]{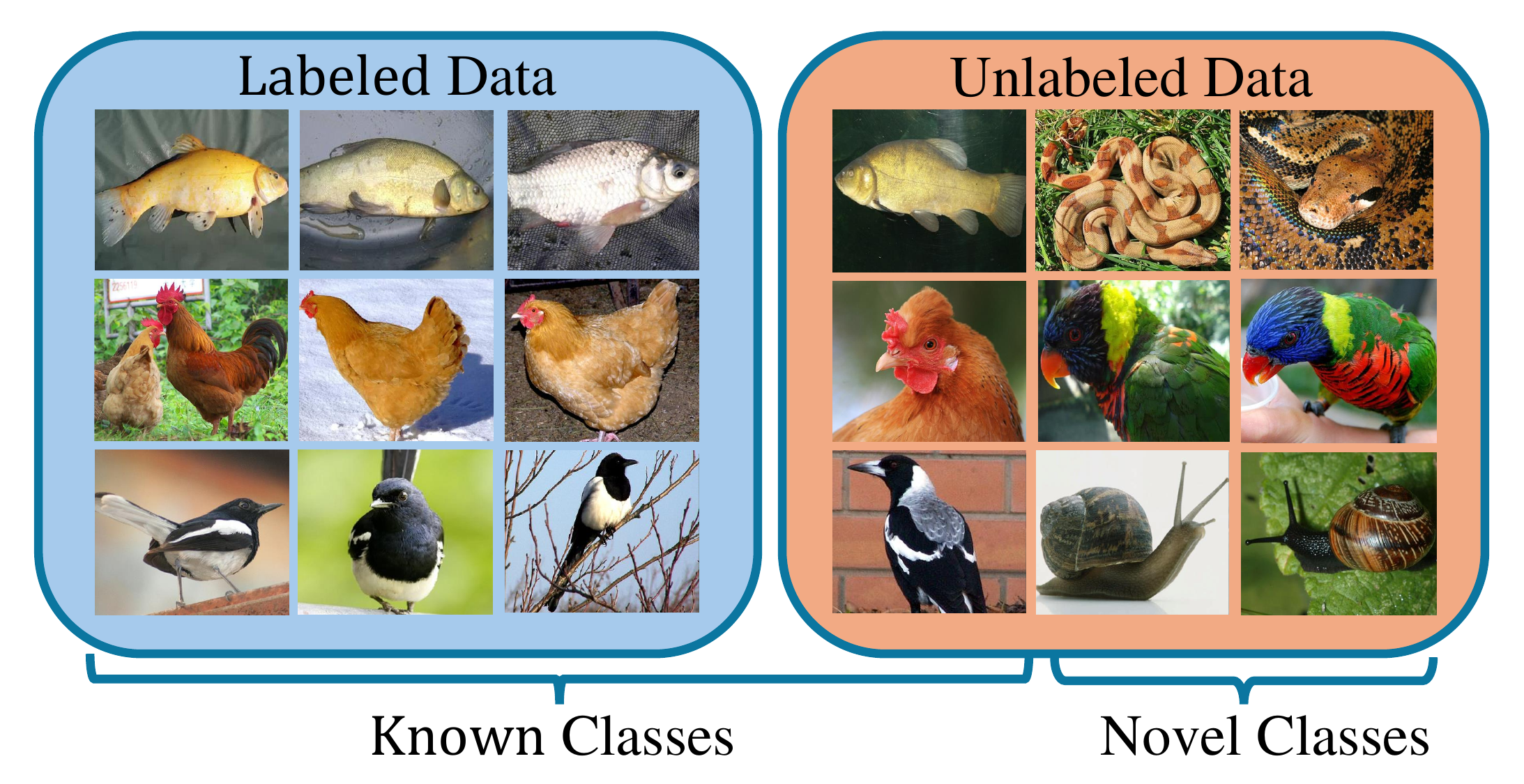} 
\caption{Generalized Category Discovery (GCD). Unlike traditional settings, GCD assumes the unlabeled data is a mixture of known and novel classes, which poses a core challenge for model learning.}
\label{gcd}
\end{figure}

Unlike New Category Discovery (NCD), Generalized Category Discovery (GCD)~\cite{vaze2022generalized} addresses a more realistic and challenging scenario where the unlabeled data contains a mixture of both known and novel classes (see Figure~\ref{gcd}). This setup renders the GCD problem more aligned with open-world settings and inherently more complex. Existing GCD methods predominantly tackle this problem from a task-specific perspective and can be broadly divided into two categories. The first category employs end-to-end pseudo-labeling strategies~\cite{han2021autonovel, fini2021unified}, which enforce class separation by assigning hard labels to unlabeled data. However, this approach is susceptible to confirmation bias, where the model can overfit to its own initial, potentially erroneous, predictions.

To mitigate this issue, a second line of work abandons pseudo-labeling in favor of a two-stage paradigm: "representation learning + clustering." Early methods in this vein often relied on non-parametric approaches, combining a self-supervised backbone with semi-supervised k-means clustering to partition the data~\cite{fei2022xcon}. While effective at alleviating overfitting on class-imbalanced datasets, these methods require executing the clustering algorithm at inference time, leading to significant computational overhead and making them unsuitable for instance-level prediction. To overcome these limitations, recent research has shifted towards parametric classifiers. For instance, SimGCD~\cite{wen2023parametric} replaced the clustering step with a classification head, enabling faster inference and establishing a strong baseline. Subsequent works, such as SPTNet~\cite{wang2024sptnet} with its spatial prompts and CMS~\cite{choi2024contrastive} with a hierarchical classification head, have further refined feature learning. However, we contend that these methods, by addressing the problem implicitly from a classification standpoint, offer symptomatic treatments rather than a fundamental solution. They lack a principled mechanism to distinguish between category-relevant signals and instance-specific noise, and thus cannot explicitly disentangle the features that define a class from the noise that hinders its discovery.

In this paper, we re-examine and address the GCD problem from the perspective of sculpting the information space. We posit that the complex, open-world nature of GCD accentuates a fundamental trade-off between the \textit{compressiveness} and \textit{informativeness} of learned representations. From an information theory \cite{cover1999elements} viewpoint, this trade-off necessitates a model that learns a representation space where, on one hand, information is effectively compressed by minimizing its dependence on the input (conditioned on the class label), and on the other hand, sufficient instance-level detail is retained to discover true novel categories. To learn such robust and minimal representations, we require a formal information-theoretic principle to explicitly govern this trade-off. Conditional Mutual Information (CMI)~\cite{yang2024conditional} has demonstrated significant effectiveness in sculpting the feature space for specific tasks. Minimizing CMI acts as a regularizer to enhance semantic consistency by pulling same-class samples closer, which improves information compression performance~\cite{yang2025conditional} and effectively addresses challenges such as class imbalance and instance-specific noise~\cite{hamidi2024fed}.

Building on this insight, this paper introduces \textbf{InfoSculpt}, which models the GCD problem by minimizing a CMI objective. InfoSculpt operationalizes this information-theoretic minimization through a dual-component framework. First, for labeled data, a \textbf{Category-Level CMI} objective forces each sample's representation to discard instance-specific details and conform to the abstract, average representation of its assigned category. This achieves principled compression by explicitly minimizing the information tied to the specific input, conditioned on its class. Second, to extend this principle to all data (especially unlabeled data) where class labels are absent, we derive a complementary \textbf{Instance-Level CMI} objective. The key insight here is to shift the conditioning context from the class label to the identity of each individual sample. This reframes the objective as minimizing the CMI between an augmented view of an input and its core instance identity. The practical effect is to train the model to distill the invariant essence of a sample while compressing the superficial information introduced by the augmentation process itself.

These two forms of CMI operate at different scales to synergistically sculpt the latent space. The Category-Level CMI leverages the strong supervisory signal from labels to shape the coarse-grained structure of the space, ensuring that known category representations are compact and well-separated. Concurrently, the Instance-Level CMI operates on all data to refine the fine-grained structure, ensuring that the representation for every single sample is a clean, stable, and robust building block, purified of noisy variations. This CMI-based objective provides the principled and systematic mechanism that prior works have lacked. Through this synergistic optimization, InfoSculpt preserves only the information essential for categorization, yielding a disentangled and discriminative representation that resolves the core conflict at the heart of GCD.

Main contributions are as follows:

\begin{itemize}
    \item \textbf{A Novel Information-Theoretic Paradigm for GCD.} We are the first to re-frame the GCD problem from an information-theoretic perspective. We model its core challenge as a fundamental trade-off between representation \textit{compressiveness} and \textit{informativeness}, providing a principled mechanism to explicitly disentangle class-defining signals from instance-specific noise, which addresses a core limitation of prior methods.

    \item \textbf{The InfoSculpt Framework with a Dual-CMI Objective.} We propose \textbf{InfoSculpt}, a novel framework that operationalizes our theory through a unique dual CMI minimization objective. It synergistically combines a \textbf{Category-Level CMI} on labeled data to enforce class compactness and a \textbf{Instance-Level CMI} on all data to distill robust, invariant instance features. Dual CMI objectives work in tandem at different granularities to systematically sculpt a high-quality latent space.

    \item \textbf{State-of-the-Art Performance.} Through comprehensive experiments on 8 benchmarks, InfoSculpt demonstrate significant performance gains over 10 existing approaches, validating its superiority.
\end{itemize}

\begin{figure*}
\centering
\includegraphics[width=0.99\textwidth]{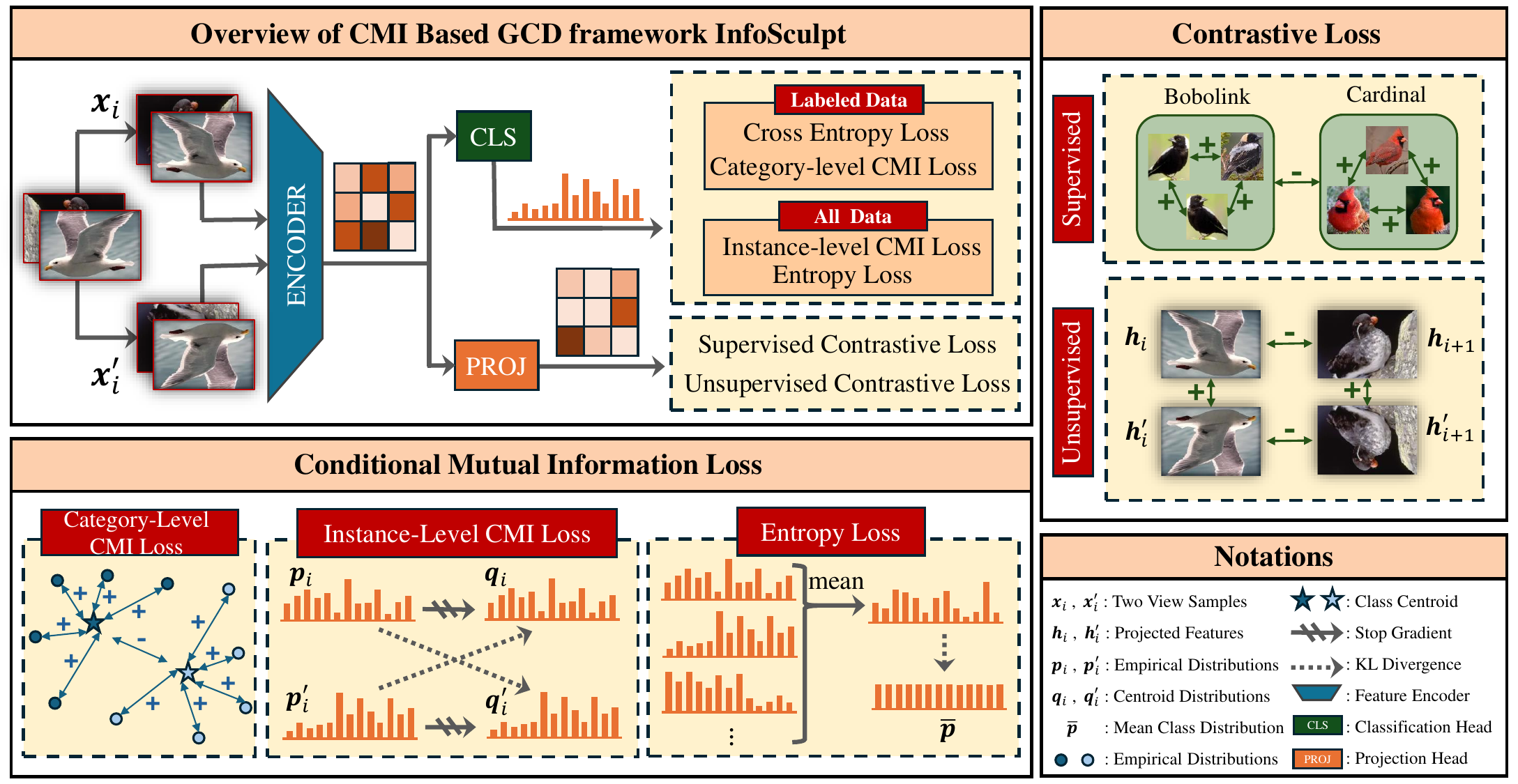} 
\caption{Architecture of InfoSculpt, which trains a feature encoder using a dual-level CMI loss—enhancing class separability and instance invariance—combined with contrastive objectives to sculpt a structured latent space tailored for GCD.
}
\label{ppl}
\end{figure*}

\section{Methodology}

\subsection{Baseline Problem Formulation for GCD.}

Formally, we consider a partially-labeled dataset $\mathcal{D} = \mathcal{D}_l \cup \mathcal{D}_u$. The labeled subset, denoted by $\mathcal{D}_l = \{(\boldsymbol{x}_i^l, y_i^l)\}_{i=1}^n \subset \mathcal{X}_l \times \mathcal{Y}_l$, comprises samples from a set of known, classes $\mathcal{C}_{\text{old}}$ (i.e., $\mathcal{Y}_l = \mathcal{C}_{\text{old}}$). The unlabeled subset, $\mathcal{D}_u = \{(\boldsymbol{x}_j^u)\}_{j=1}^m \subset \mathcal{X}_u$, contains samples whose underlying label space $\mathcal{Y}_u$ spans both the old classes $\mathcal{C}_{\text{old}}$ and a set of novel classes $\mathcal{C}_{\text{new}}$, such that $\mathcal{Y}_u = \mathcal{C}_{\text{old}} \cup \mathcal{C}_{\text{new}}$.

The objective of GCD is to leverage the supervision from $\mathcal{D}_l$ to simultaneously cluster the novel-class samples from $\mathcal{C}_{\text{new}}$ and classify the old-class samples from $\mathcal{C}_{\text{old}}$ that are present in $\mathcal{D}_u$. The number of old classes, $K_{\text{old}} = |\mathcal{C}_{\text{old}}|$, is directly inferred from $\mathcal{D}_l$. The number of novel classes $K_{\text{new}} = |\mathcal{C}_{\text{new}}|$, is assumed to be known a priori. The total number of classes is $K = K_{\text{old}} + K_{\text{new}}$.

As applying contrastive loss directly on classification features would cause conflicting objectives \cite{chen2020simple}, InfoSculpt uses dual projection heads to decouple the two feature spaces. Let $\mathcal{E}(\cdot)$ denote the feature extractor and $\phi(\cdot)$ be the projection head. For a given input sample $\boldsymbol{x}_i$, its $d$-dimensional feature representation is $\boldsymbol{z}_i = \mathcal{E}(\boldsymbol{x}_i)$. This representation is subsequently mapped to a $d_h$-dimensional projection space for contrastive learning via $\boldsymbol{h}_i = \phi(\boldsymbol{z}_i)$. In parallel, a classifier head, denoted by $\psi(\cdot)$, maps the same feature representation $\boldsymbol{z}_i$ to the $K$-dimensional logits $\boldsymbol{p}_i = \psi(\boldsymbol{z}_i)$, which are used for the final class prediction.

\subsection{Formulate GCD from Information Theory}

\subsubsection{Modeling GCD From IB to a Tractable CMI Objective}

\begin{figure}
\centering
\includegraphics[width=0.99\columnwidth]{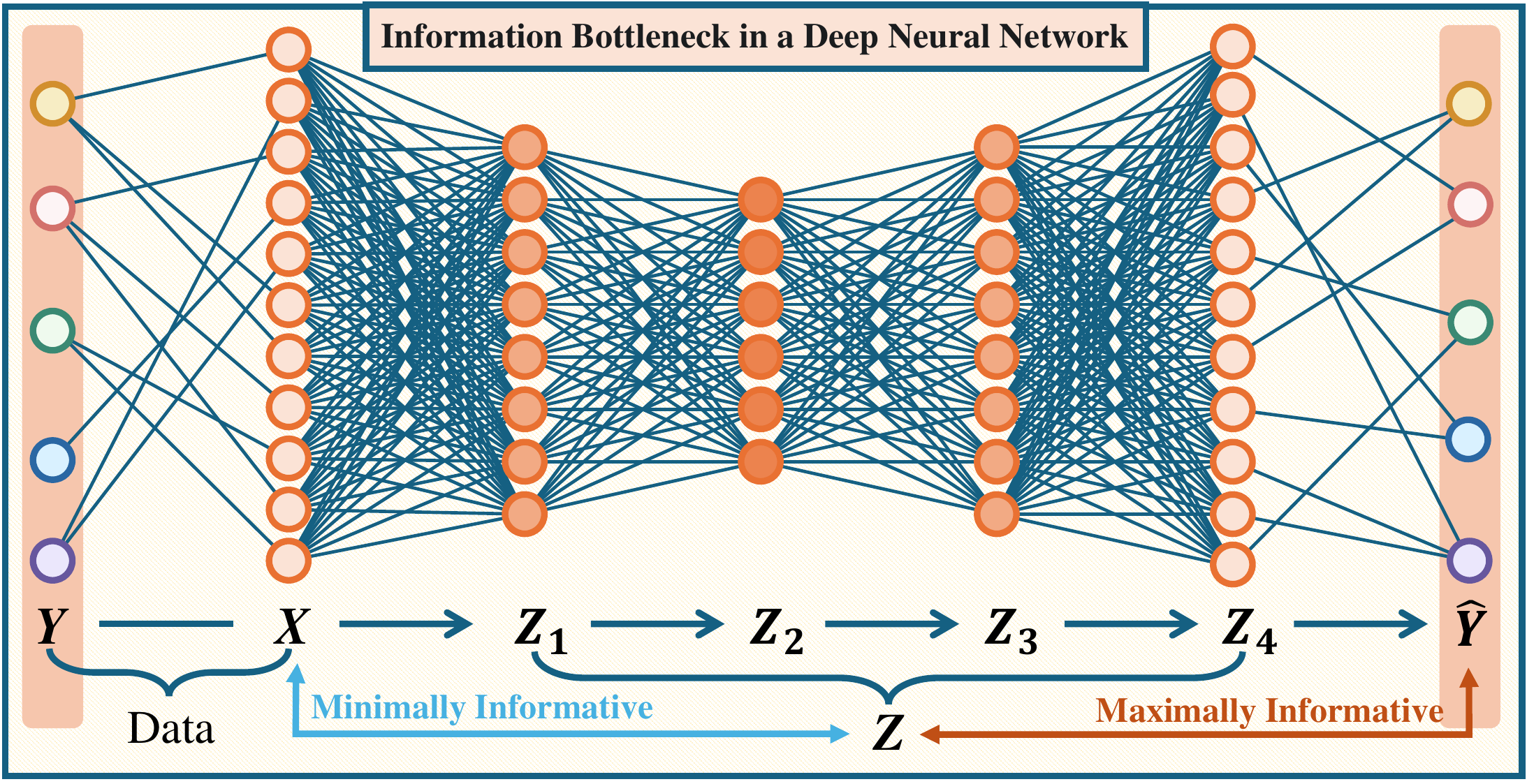} 
\caption{Illustration of the IB principle. The network learns a compressed internal representation, \textit{Z} that is maximally informative about the true label \textit{Y} while discarding irrelevant information from \textit{X}.}
\label{cmi}
\end{figure}

We begin by reformulating the GCD trade-off through the lens of the the Information Bottleneck (IB) principle~\cite{tishby2000information, tishby2015deep}. As illustrated in Figure~\ref{cmi}, the goal is to learn a representation $Z$ from an input $X$ that perfectly balances two objectives. First, it must be \textbf{minimal}, meaning it is maximally compressed by discarding all input information irrelevant to the class label $Y$. Second, it must be \textbf{sufficient}, meaning it remains maximally informative about $Y$ to ensure predictive accuracy and generalization. This fundamental trade-off is mathematically formalized by the IB Lagrangian \cite{tishby2000information}:

\begin{equation}
    \mathcal{L}_{\text{IB}} = I(X; Z) - \beta I(Y; Z),
\end{equation}

\noindent where $I(\cdot; \cdot)$ is the mutual information. The term $I(X; Z)$ serves as the compression cost to be minimized, while $I(Y; Z)$ represents the predictive value to be maximized. The hyperparameter $\beta$ explicitly controls the trade-off between these two goals, to find an optimal balance between a representation that is discriminative yet maximally compact.

However, the direct application of IB to modern neural networks faces two major challenges. First, for deterministic networks, the representation $Z$ is a function of the input $X$, causing $I(X; Z)$ to degenerate into the entropy $H(Z)$, which weakens the original theoretical motivation. Second, estimating mutual information for high-dimensional and continuous variables like $X$ and $Z$ is notoriously difficult and often inaccurate.

Inspired by \cite{zhuang2025stealthy}, a specialized and tractable form of this principle is adopted. We consider the model's output logits, denoted by the random variable $\hat{Y}$, as the compressed representation (i.e., $Z = \hat{Y}$). This leads to the CMI, $I(X; \hat{Y}|Y)$:

\begin{equation}
    I(X; \hat{Y}|Y) = I(X; \hat{Y}) - I(Y; \hat{Y}).
\end{equation}

It can be shown that minimizing $I(X; \hat{Y}|Y)$ is equivalent to a specific form of the IB objective. Intuitively, minimizing $I(X; \hat{Y}|Y)$ aims to create a prediction $\hat{Y}$ that is highly informative about the true label $Y$ (maximizing $I(Y; \hat{Y})$) while being minimally informative about the specific input instance $X$ (minimizing $I(X; \hat{Y})$). This encourages the model to learn class-level features rather than memorizing instance-specific noise.

\subsubsection{A Practical Estimator.}

To develop a practical metric, we derive a computable estimator for $I(X; \hat{Y}|Y)$. We begin from a standard expansion of the CMI:
\begin{equation}
    I(X; \hat{Y}|Y) = \sum_{x, y, \hat{y}} P(x, y, \hat{y}) \log \frac{P(\hat{y}|x, y)}{P(\hat{y}|y)}.
    \label{eq:cmi_expanded}
\end{equation}

The derivation now pivots on a key assumption that models the information flow within the standard supervised learning paradigm. A neural network classifier is trained to generate its prediction $\hat{Y}$ based solely on the input $X$, without direct access to the true label $Y$. This operational principle is formally captured by the \textbf{Markov chain $Y \rightarrow X \rightarrow \hat{Y}$}. This chain asserts that the prediction $\hat{Y}$ is conditionally independent of the true label $Y$ given the input $X$, which implies the simplification $P(\hat{y}|x, y) = P(\hat{y}|x)$. Substituting this conditional independence property into Equation~\ref{eq:cmi_expanded} directly simplifies the expression:
\begin{align}
    I(X; \hat{Y}|Y) &= \sum_{x, y, \hat{y}} P(x, y, \hat{y}) \log \frac{P(\hat{y}|x)}{P(\hat{y}|y)} \\
    &= \sum_{x,y} P(x,y) \sum_{\hat{y}} P(\hat{y}|x) \log \frac{P(\hat{y}|x)}{P(\hat{y}|y)} \label{eq:markov_applied} \\
    &= \mathbb{E}_{(X,Y)} \left[ \underbrace{D_{\text{KL}}\big( P(\hat{Y}|X=x) \parallel P(\hat{Y}|Y=y) \big)}_{\text{KL Divergence}} \right]. \label{eq:final_kl}
\end{align}

The inner summation is the definition of the Kullback-Leibler (KL) divergence between the distributions $P(\hat{Y}|X=x)$ and $P(\hat{Y}|Y=y)$. This allows to express the CMI in its more intuitive and compact form:

\begin{equation} \label{eq:cmi_theoretical_kl}
    I(X; \hat{Y}|Y) = \mathbb{E}_{(X,Y)}\left[ \mathrm{KL}\left( P(\hat{Y}|X) \,\|\, P(\hat{Y}|Y) \right) \right].
\end{equation}

To operationalize this, we find empirical estimators for the two distributions using labeled data $\mathcal{D}_l = \{(\boldsymbol{x}_i, y_i)\}_{i=1}^n$. 

The instance-specific posterior is directly given by the model's softmax output for input $\boldsymbol{x}_i$. This $K$-dimensional vector is denoted as $\boldsymbol{p}_i$:

\begin{equation}
    P(\hat{Y}=\hat{y}|X=\boldsymbol{x}_i) := [\boldsymbol{p}_i]_{\hat{y}} = [\text{softmax}(\psi(\mathcal{E}(\boldsymbol{x}_i)))]_{\hat{y}}.
\end{equation}

The class-conditional posterior is the model's expected prediction for class $y$. Using the law of total expectation and the Markov chain $Y \rightarrow X \rightarrow \hat{Y}$:
\begin{align}
    P(\hat{Y}=\hat{y}|Y=y) &= \sum_{x \in \mathcal{X}} P(x|y) P(\hat{y}|x,y) \\&= \sum_{x \in \mathcal{X}} P(x|y) P(\hat{y}|x) \\
    &= \mathbb{E}_{X|Y=y}{P(\hat{Y}=\hat{y}|X)}.
\end{align}

We estimate this expectation by taking the sample mean of the posteriors of all instances in $\mathcal{D}_l$ with the true label $y$. This defines the empirical class centroid vector $\boldsymbol{q}^y$:

\begin{equation}
    [\boldsymbol{q}^y]_{\hat{y}} \approx \frac{1}{|\mathcal{D}_y|} \sum_{i: (\boldsymbol{x}_i, y_i) \in \mathcal{D}_y} [\boldsymbol{p}_i]_{\hat{y}},
\end{equation}
where $\mathcal{D}_y = \{(\boldsymbol{x}_i, y_i) \in \mathcal{D}_l \mid y_i=y\}$.

\paragraph{The Empirical CMI Estimator.} Finally, we approximate the outer expectation in Equation~\eqref{eq:cmi_theoretical_kl} with a sample mean over the $n$ samples in $\mathcal{D}_l$. Substituting the empirical distributions $\boldsymbol{p}_i$ and $\boldsymbol{q}^{y_i}$ yields the practical CMI estimator:

\begin{equation} \label{eq:cmi_estimator_final}
    I(X; \hat{Y}|Y) = \frac{1}{n} \sum_{i=1}^{n} \mathrm{KL}\left( \boldsymbol{p}_i \,\|\, \boldsymbol{q}^{y_i} \right).
\end{equation}

This expression provides a robust and computable metric under the information-theoretic framework, offering deep insight into the model's predictive structure.

\subsection{Estimating GCD via CMI}

As shown in Figure \ref{ppl}, the final training objective is a composite loss function that strategically combines different terms to leverage both labeled and unlabeled data. The components are designed to ensure accurate supervision, promote intra-class compactness, enforce instance-level consistency, and prevent model collapse.

For the labeled data $\mathcal{D}_l$, the objective combines two terms. First, the standard supervised cross-entropy loss is used to learn from ground-truth labels:

\begin{equation}
\mathcal{L}_{\text{ce}} = \frac{1}{|\mathcal{B}_l|} \sum_{i \in \mathcal{B}_l} \ell(\mathbf{y}_i, \boldsymbol{p}_i),
\end{equation}

\noindent where $\mathcal{B}_l$ is a mini-batch of labeled samples and $\mathbf{y}_i$ is the one-hot ground-truth label. 

Second, to learn a compressed representation, we incorporate the CMI estimator from Equation~\eqref{eq:cmi_estimator_final} as a regularization term. To create a target that sharpens class focus while mitigating overfitting, we refine the empirical class-conditional distribution $\boldsymbol{q}^{y_i}$. Specifically, we construct a modified target, $\boldsymbol{\hat{q}}^{y_i}$, from $\boldsymbol{q}^{y_i}$ by: (i) setting the ground-truth class confidence to 1, (ii) suppressing the top-$k$ non-ground-truth confidences to 0, and (iii) retaining the remaining low-confidence values. This procedure sharpens focus by maximizing the ground-truth signal and eliminating the most distracting hard-negative classes. Simultaneously, retaining the original low-confidence values preserves soft, data-driven knowledge about inter-class similarities, acting as a regularizer to prevent the model from making overconfident negative predictions. The final loss term is the KL divergence between the prediction and this refined target:

\begin{equation}
\mathcal{L}_{\text{cmi}} = \frac{1}{|\mathcal{B}_l|} \sum_{i \in \mathcal{B}_l} \mathrm{KL}\left( \boldsymbol{p}_i \,\|\, \boldsymbol{\hat{q}}^{y_i} \right).
\label{eq:refined_cat_cmi}
\end{equation}

To ensure that the learned class centers are distinct and well-separated in the embedding space, a separation loss is introduced:

\begin{equation}
\mathcal{L}_{\text{sep}} = \frac{1}{K} \sum_{i=1}^{K} \log \frac{1}{K-1} \sum_{j=1, j \neq i}^{K} \exp\left( (\boldsymbol{q}^{y_i})^{\top} \boldsymbol{q}^{y_j} / \tau_{\text{sep}} \right),
\end{equation}

\noindent where $K$ is the total number of the classes and $\tau_{\text{sep}}$ is a temperature that controls the sensitivity to this similarity. To effectively utilize the entire training set $\mathcal{D}$, including unlabeled data, the information-theoretic perspective is extended to all samples. 
We introduce an instance-level CMI objective, which re-frames cross-view consistency as a CMI minimization task. In this view, each sample is treated as its own unique class. The sharpened prediction from one augmented view, $\boldsymbol{q}_i'$, serves as the target distribution for the prediction from the other view, $\boldsymbol{p}_i$. This objective is applied symmetrically, where the sharpened prediction $\boldsymbol{q}_i$ from the first view also serves as the target for the second view's prediction, $\boldsymbol{p}_i'$. This symmetric loss, applied over a mini-batch $\mathcal{B}$ from the entire dataset, is formulated as:

\begin{equation}
\mathcal{L}_{\text{inst}} = \frac{1}{2|\mathcal{B}|} \sum_{i \in \mathcal{B}} \left( \mathrm{KL}(\boldsymbol{q}_i' \,\|\, \boldsymbol{p}_i) + \mathrm{KL}(\boldsymbol{q}_i \,\|\, \boldsymbol{p}_i') \right).
\end{equation}

Finally, to prevent the model from collapsing to trivial solutions by assigning all samples to a few clusters, we maximize the entropy of the marginal predictions over the batch. This encourages diverse and balanced cluster assignments:

\begin{equation}
\mathcal{L}_{\text{ent}} = -H(\bar{\boldsymbol{p}}) = \sum_{k=1}^{K} \bar{\boldsymbol{p}}^{(k)} \log \bar{\boldsymbol{p}}^{(k)},
\end{equation}
where $\bar{\boldsymbol{p}} = \frac{1}{2|\mathcal{B}|} \sum_{i \in \mathcal{B}} (\boldsymbol{p}_i + \boldsymbol{p}_i')$.

The cluster training loss is a weighted sum of these four components:
\begin{equation}
\mathcal{L_{\text{cls}}} = \lambda_{\text{cmi}}\mathcal{L}_{\text{cmi}} + \lambda_{\text{sep}}\mathcal{L}_{\text{sep}} + \lambda_{\text{inst}}\mathcal{L}_{\text{inst}} + \lambda_{\text{ent}}\mathcal{L}_{\text{ent}},
\end{equation}

\noindent where $\lambda_{\text{cmi}}$, $\lambda_{\text{sep}}$, $\lambda_{\text{inst}}$, and $\lambda_{\text{ent}}$ are hyperparameters that balance the influence of each regularization term.

\subsection{Representation Learning}
The representation learning process in the framework follows the protocols of GCD \cite{vaze2022generalized} and SimGCD \cite{wen2023parametric}. Specifically, we employ a Vision Transformer (ViT-B/16)~\cite{dosovitskiy2020image} pretrained using DINO self-supervised learning~\cite{caron2021emerging} on the ImageNet dataset~\cite{russakovsky2015imagenet} as the backbone network. The learning process consists of two stages: supervised contrastive learning applied to $\mathcal{D}_l$, and unsupervised contrastive learning conducted on $\mathcal{D}$.

Specifically, the supervised contrastive learning \cite{khosla2020supervised} on labeled data is:
\begin{equation} \label{eq:supervised_loss}
\mathcal{L}_{\text{con}}^{l} = \frac{1}{|\mathcal{B}_l|} \sum_{i \in \mathcal{B}_l} \frac{1}{|\mathcal{N}(i)|} \sum_{q \in \mathcal{N}(i)} -\log \frac{\exp(\boldsymbol{h}_i^\top \boldsymbol{h}_q / \tau_c)}{\sum_{j} \mathbb{1}_{[j \neq i]} \exp(\boldsymbol{h}_i^\top \boldsymbol{h}_j / \tau_c)},
\end{equation}

\noindent where $\mathcal{N}(i)$ denotes positive samples with the same label as $\boldsymbol{x}_i$. The $\mathbb{1}_{[\cdot]}$ denotes the indicator function and equals to 1 when the condition is true else 0, $\tau_c$ denotes the temperature in contrastive learning.

Given two views (random augmentations) of the input $\boldsymbol{x}_i$ and $\boldsymbol{x}'_i$ in a mini-batch $\mathcal{B}$, the unsupervised contrastive learning loss:
\begin{equation} \label{eq:unsupervised_loss}
\mathcal{L}_{\text{con}}^{u} = \frac{1}{|\mathcal{B}|} \sum_{i \in \mathcal{B}} -\log \frac{\exp(\boldsymbol{h}_i^\top \boldsymbol{h}'_i / \tau_c)}{\sum_{j} \mathbb{1}_{[j \neq i]} \exp(\boldsymbol{h}_i^\top \boldsymbol{h}_j / \tau_c)}.
\end{equation}

By integrating the learning objectives above, we could obtain the overall learning objective:

\begin{equation} \label{eq:all}
\mathcal{L} =\alpha\mathcal{L}_{\text{ce}} + (1 - \alpha)\mathcal{L}_{\text{cls}} + \beta\mathcal{L}_{\text{con}}^{l} + (1 - \beta)\mathcal{L}_{\text{con}}^{u},
\end{equation}

\noindent where $\alpha$ and $\beta$ are balancing coefficients.

\section{Experiments}
\label{sec:experiments}

\subsection{Datasets and Evaluation Protocol}

\paragraph{Datasets.}
The effectiveness of InfoSculpt is validated on 8 benchmark datasets, consistent with prior work like SimGCD, including CIFAR-10/100~\cite{krizhevsky2009learning}, ImageNet-100~\cite{tian2020contrastive}, as well as fine-grained classification datasets CUB-200-2011~\cite{wah2011caltech}, Stanford Cars~\cite{krause20133d}, and FGVC-Aircraft~\cite{maji2013fine}. To further assess the robustness of InfoSculpt, we include two more challenging large-scale datasets: Herbarium 19~\cite{tan2019herbarium} and the full ImageNet-1K~\cite{russakovsky2015imagenet}.

Each dataset is adhered to the standard GCD protocol established in SimGCD. Specifically, we form the labeled training set, $\mathcal{D}^l$, by sub-sampling 50\% of the images from the known classes. The remaining images from these known classes, combined with all images from the novel classes, constitute the unlabeled set, $\mathcal{D}^u$. A detailed overview of the dataset splits is provided in Table~\ref{tab:datasets}.

\begin{table}
\centering
\begin{tabular}{l|rc|rc}
\toprule
\multirow{2}{*}{\textbf{Dataset}} & \multicolumn{2}{c|}{\textbf{Labeled $\mathcal{D}^l$}} & \multicolumn{2}{c}{\textbf{Unlabeled $\mathcal{D}^u$}} \\ \cline{2-5} 
& \textbf{Images} & \textbf{Old} & \textbf{Images} & \textbf{New} \\ 
\midrule
CIFAR-10& 12.5k & 5  & 37.5k & 10 \\
CIFAR-100& 20.0k & 80 & 30.0k & 100 \\
ImageNet-100& 31.9k & 50 & 95.3k & 100 \\
CUB-200-2011& 1.5k  & 100& 4.5k  & 200 \\
Stanford Cars& 2.0k  & 98 & 6.1k  & 196 \\
FGVC-Aircraft & 1.7k  & 50 & 5.0k  & 100 \\
Herbarium 19& 8.9k  & 341& 25.4k & 683 \\
ImageNet-1K& 321k  & 500& 960k  & 1000 \\
\bottomrule
\end{tabular}
\caption{Overview of the datasets used in the experiments. We list the specific number of labeled and unlabeled images ($\mathcal{D}^l$, $\mathcal{D}^u$) and their corresponding class assignments (“Old” and “New”).}
\label{tab:datasets}
\end{table}

\paragraph{Evaluation Protocol.}
During training, the entire dataset $\mathcal{D} = \mathcal{D}^l \cup \mathcal{D}^u$ is utilized. For evaluation, InfoSculpt measure performance using the standard unsupervised clustering accuracy (ACC) metric~\cite{vaze2022generalized}, computed on all samples in the unlabeled set $\mathcal{D}^u$. Given the ground-truth labels $y_i^*$ and the model's predicted cluster assignments $\hat{y}_i$, ACC is defined as:
\begin{equation}
\label{eq:acc}
\text{ACC} = \frac{1}{M} \sum_{i=1}^{M} \mathbb{1}\left(y_i^* = \text{map}(\hat{y}_i)\right),
\end{equation}
where $M = |\mathcal{D}^u|$ is the total number of unlabeled samples, $\mathbb{1}(\cdot)$ is the indicator function, and $\text{map}(\cdot)$ is the optimal permutation mapping function. This mapping is determined using the Hungarian algorithm~\cite{kuhn1955hungarian} to find the best one-to-one alignment between the predicted cluster indices and the ground-truth class labels.

\begin{table*}
\caption{Comparison of InfoSculpt with SOTA approaches on CIFAR-10, CIFAR-100, and ImageNet-100.}
\label{tab:main_results_4}
\centering
\small
\resizebox{\textwidth}{!}{%
\begin{tabular}{l|ccc|ccc|ccc|ccc}
\toprule
\multirow{2}{*}{\textbf{Models}} & \multicolumn{3}{c|}{\textbf{CIFAR10}} & \multicolumn{3}{c|}{\textbf{CIFAR100}} & \multicolumn{3}{c|}{\textbf{ImageNet-100}} & \multicolumn{3}{c}{\textbf{ImageNet-1k}} \\
\cmidrule(lr){2-4} \cmidrule(lr){5-7} \cmidrule(lr){8-10} \cmidrule(lr){11-13}
 & All & Old & New & All & Old & New & All & Old & New & All & Old & New \\
\midrule
GCD (CVPR'22)& 91.5 & 97.9 & 88.2 & 73.0 & 76.2 & 66.5 & 74.1 & 89.8 & 66.3 & 52.5 & 72.5 & 42.2 \\
DCCL (CVPR'23)& 96.3 & 96.5 & 96.9 & 75.3 & 76.8 & 70.2 & 80.5 & 90.5 & 76.2 & - & - & - \\
GPC (ICCV'23)& 92.2 & 98.2 & 89.1 & 77.9 & 85.0 & 63.0 & 76.9 & 94.3 & 71.0 & - & - & - \\
SimGCD (ICCV'23)& 97.1 & 95.1 & 98.1 & 80.1 & 81.2 & 77.8 & 83.0 & 93.1 & 77.9 & 57.1 & 77.3 & 46.9 \\
LegoGCD (CVPR'24) & 97.1 & 94.3 & 98.5 & 81.8 & 81.4 & 82.5 & 86.3 & 94.5 & 82.1 & 62.4 & 79.5 & 53.8 \\
ActiveGCD (CVPR'24) & 93.2 & 94.6 & 92.8 & 71.3 & 75.7 & 66.8 & 83.3 & 90.2 & 76.5 & - & - & - \\
MTMC (arXiv'25) & - & - & - & 80.2 & 81.5 & 77.5 & 86.7 & 93.1 & 83.6 & - & - & - \\
ProtoGCD (TPAMI'25) & 97.3 & 95.3 & 98.2 & 81.9 & 82.9 & 80.0 & 84.0 & 92.2 & 79.9 & - & - & - \\
\midrule
\rowcolor{lightblue} Ours & \textbf{97.4}{\tiny,±0.0} & {95.3}{\tiny,±0.2} & \textbf{98.6}{\tiny,±0.0} & \textbf{82.2}{\tiny,±0.2} & \textbf{83.3}{\tiny,±0.0} & {78.0}{\tiny,±0.4} & \textbf{85.6}{\tiny,±0.6} & \textbf{93.7}{\tiny,±0.9} & {77.5}{\tiny,±1.3} & \textbf{63.0}{\tiny,±0.3} & \textbf{79.7}{\tiny,±0.3} & \textbf{54.1}{\tiny,±0.5} \\
\bottomrule
\end{tabular}
}
\end{table*}

\begin{table*}
\caption{Comparison of InfoSculpt with SOTA approaches on CUB, Stanford Cars, FGVC-Aircraft and Herbarium 19.}
\label{tab:main_results_4_finegrained}
\centering
\small
\resizebox{\textwidth}{!}{%
\begin{tabular}{l|ccc|ccc|ccc|ccc}
\toprule
\multirow{2}{*}{\textbf{Models}} & \multicolumn{3}{c|}{\textbf{CUB}} & \multicolumn{3}{c|}{\textbf{Stanford Cars}} & \multicolumn{3}{c|}{\textbf{FGVC-Aircraft}} & \multicolumn{3}{c}{\textbf{Herbarium 19}} \\
\cmidrule(lr){2-4} \cmidrule(lr){5-7} \cmidrule(lr){8-10} \cmidrule(lr){11-13}
 & All & Old & New & All & Old & New & All & Old & New & All & Old & New \\
\midrule
GCD (CVPR'22)& 51.3 & 56.6 & 48.7 & 39.0 & 57.6 & 29.9 & 45.0 & 41.1 & 46.9 & 35.4 & 51.0 & 27.0 \\
DCCL (CVPR'23)& 63.5 & 60.8 & 64.9 & 43.1 & 55.7 & 36.2 & - & - & - & - & - & - \\
GPC (ICCV'23)& 55.4 & 58.2 & 53.1 & 42.8 & 59.2 & 32.8 & 46.3 & 42.5 & 47.9 & - & - & - \\
SimGCD (ICCV'23)& 60.3 & 65.6 & 57.7 & 53.8 & 71.9 & 45.0 & 54.2 & 59.1 & 51.8 & 44.0 & 58.0 & 36.4 \\
LegoGCD (CVPR'24) & 63.8 & 71.9 & 59.8 & 57.3 & 75.7 & 48.4 & 55.0 & 61.5 & 51.7 & 45.1 & 57.4 & 38.4 \\
ActiveGCD (CVPR'24) & 66.6 & 66.5 & 66.7 & 48.4 & 57.7 & 39.3 & 53.7 & 51.5 & 56.0 & - & - & - \\
MTMC (arXiv'25) & 62.1 & 65.8 & 60.3 & 52.3 & 70.0 & 43.7 & 55.1 & 58.9 & 53.1 & 45.6 & 57.8 & 39.0 \\
Hyp-GCD (CVPR'25) & 61.0 & 67.0 & 58.0 & 50.8 & 60.9 & 45.8 & 48.2 & 43.6 & 50.5 & - & - & - \\
ProtoGCD (TPAMI'25) & 65.2 & 68.5 & 60.5 & 53.8 & 73.7 & 44.2 & 56.8 & 62.5 & 53.9 & 44.5 & 59.4 & 36.5 \\
\midrule
\rowcolor{lightblue} InfoSculpt & \textbf{66.8}{\tiny,±0.1} & \textbf{72.0}{\tiny,±0.4} & {59.9}{\tiny,±0.2} & \textbf{59.5}{\tiny,±0.2} & \textbf{78.4}{\tiny,±0.4} & {42.1}{\tiny,±0.7} & \textbf{57.0}{\tiny,±0.4} & \textbf{65.0}{\tiny,±0.6} & {50.1}{\tiny,±1.0} & \textbf{48.5}{\tiny,±0.2} & \textbf{61.2}{\tiny,±0.4} & {36.0}{\tiny,±0.4} \\
\bottomrule
\end{tabular}
}
\end{table*}

\begin{figure*}[!t]
\centering
\begin{subfigure}{0.245\textwidth}
    \includegraphics[width=\linewidth]{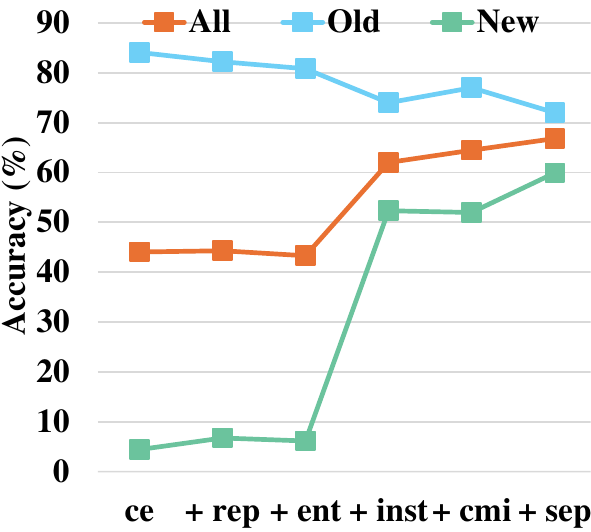}
    \caption{CUB}
\end{subfigure}
\hfill
\begin{subfigure}{0.245\textwidth}
    \includegraphics[width=\linewidth]{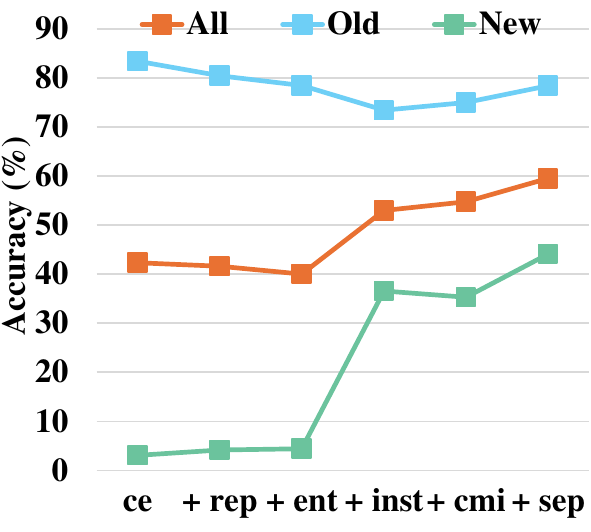}
    \caption{Stanford Cars}
\end{subfigure}
\hfill
\begin{subfigure}{0.245\textwidth}
    \includegraphics[width=\linewidth]{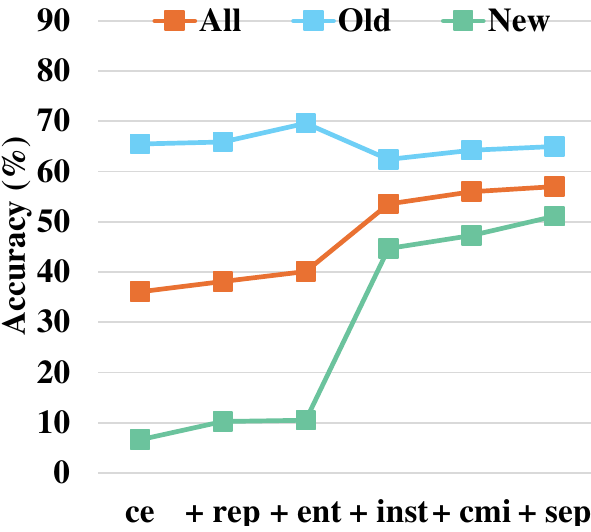}
    \caption{FGVC-Aircraft}
\end{subfigure}
\hfill
\begin{subfigure}{0.245\textwidth}
    \includegraphics[width=\linewidth]{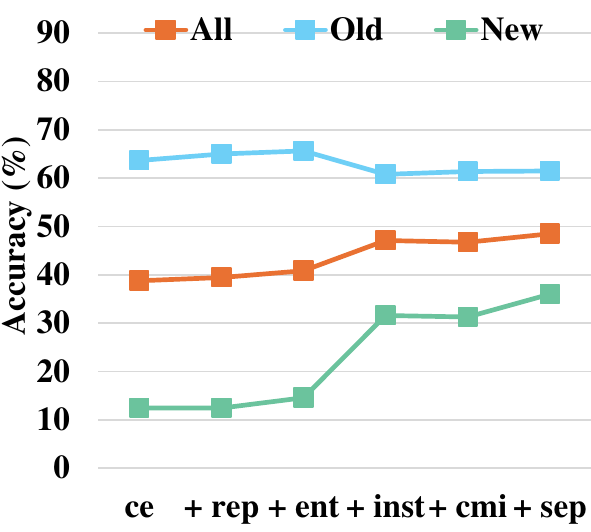}
    \caption{Herbarium 19}
\end{subfigure}
\caption{Ablation study of InfoSculpt on four fine-grained datasets.}
\label{fig:ablations}
\end{figure*}

\begin{figure*}
\centering
\begin{subfigure}{0.24\textwidth}
    \includegraphics[width=\linewidth]{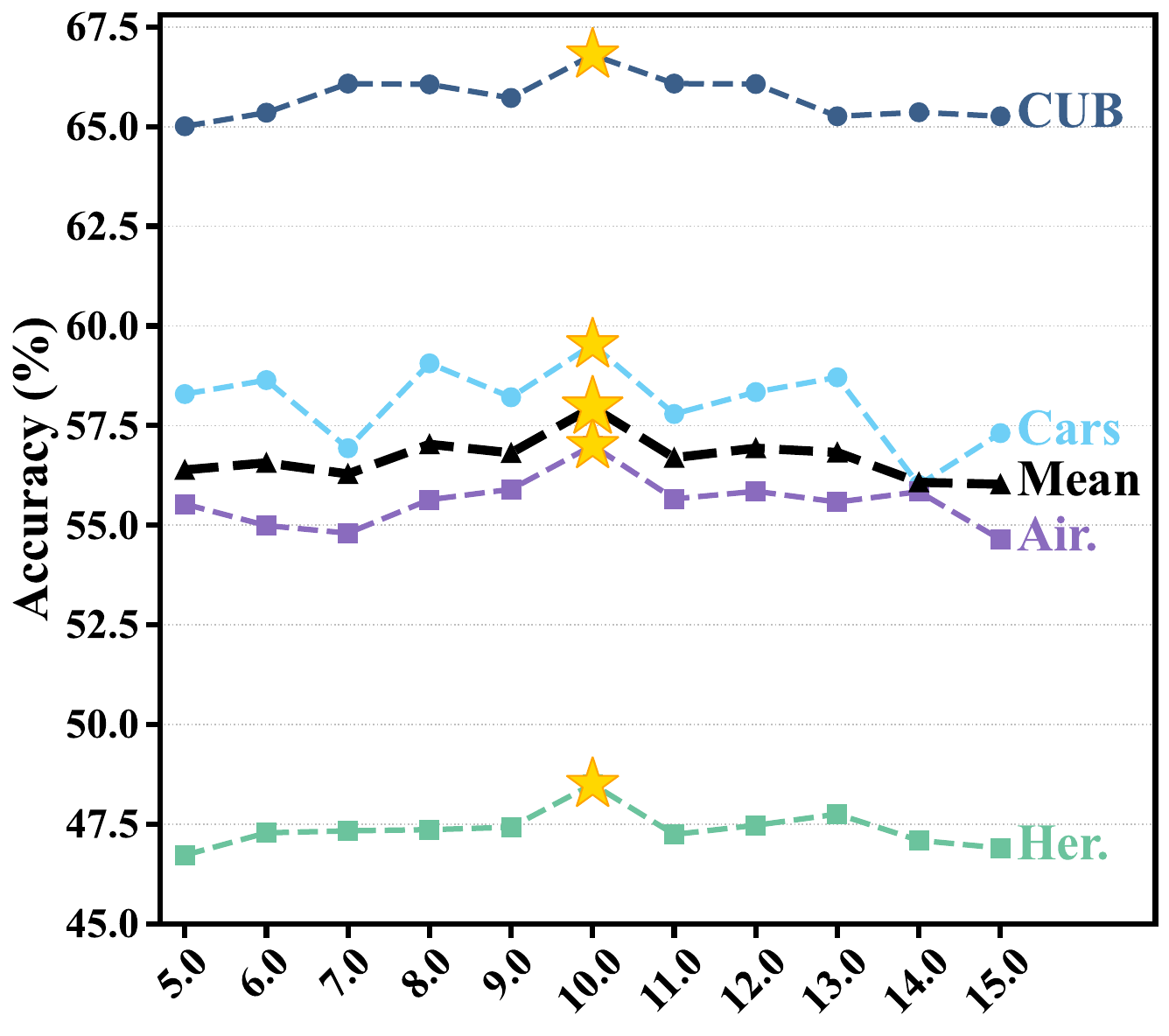}
    \caption{$k$}
    \label{subfig:k}
\end{subfigure}
\hfill
\begin{subfigure}{0.24\textwidth}
    \includegraphics[width=\linewidth]{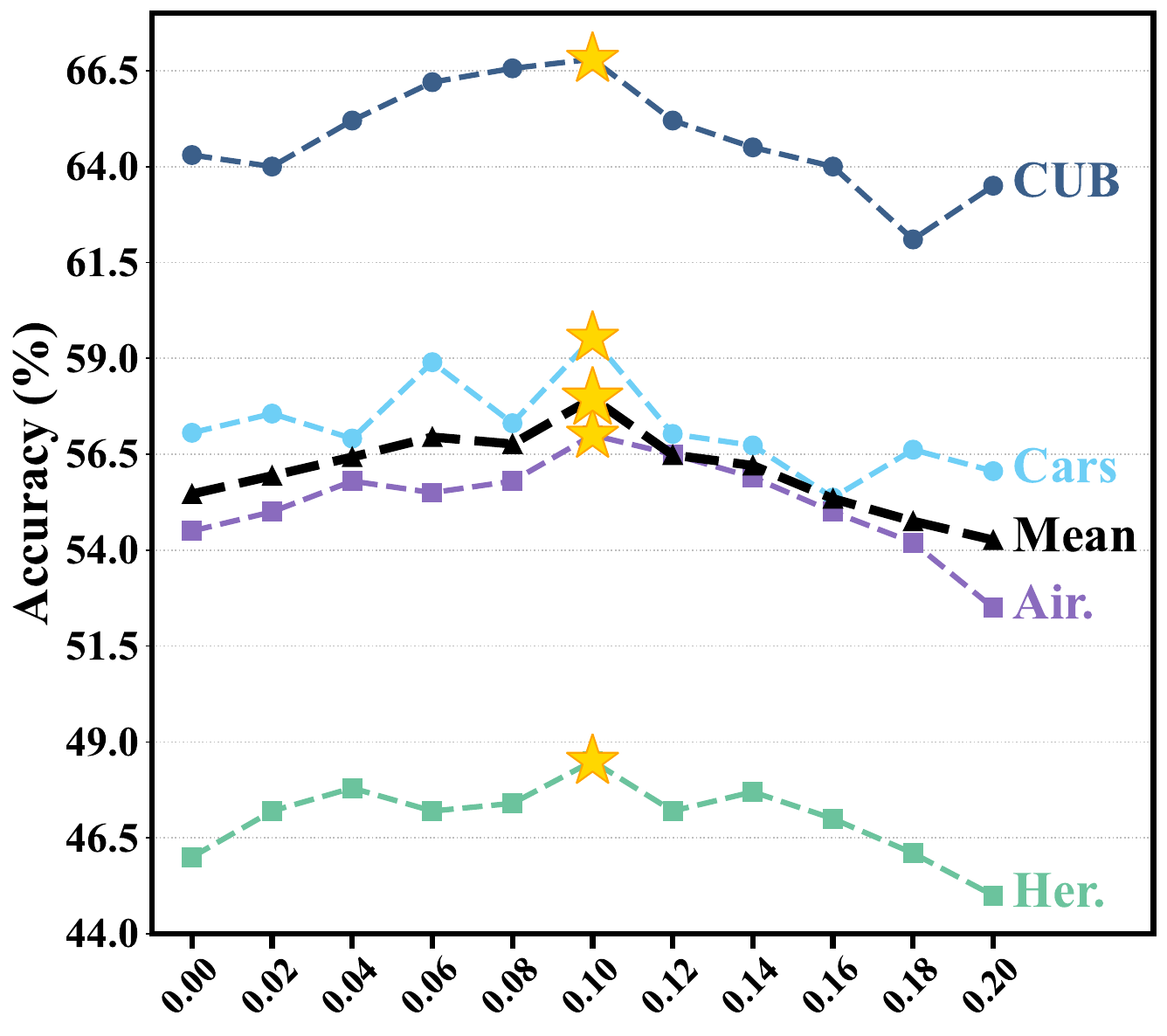}
    \caption{$\lambda_{{cmi}}$}
    \label{subfig:lambda_cmi}
\end{subfigure}
\hfill
\begin{subfigure}{0.24\textwidth}
    \includegraphics[width=\linewidth]{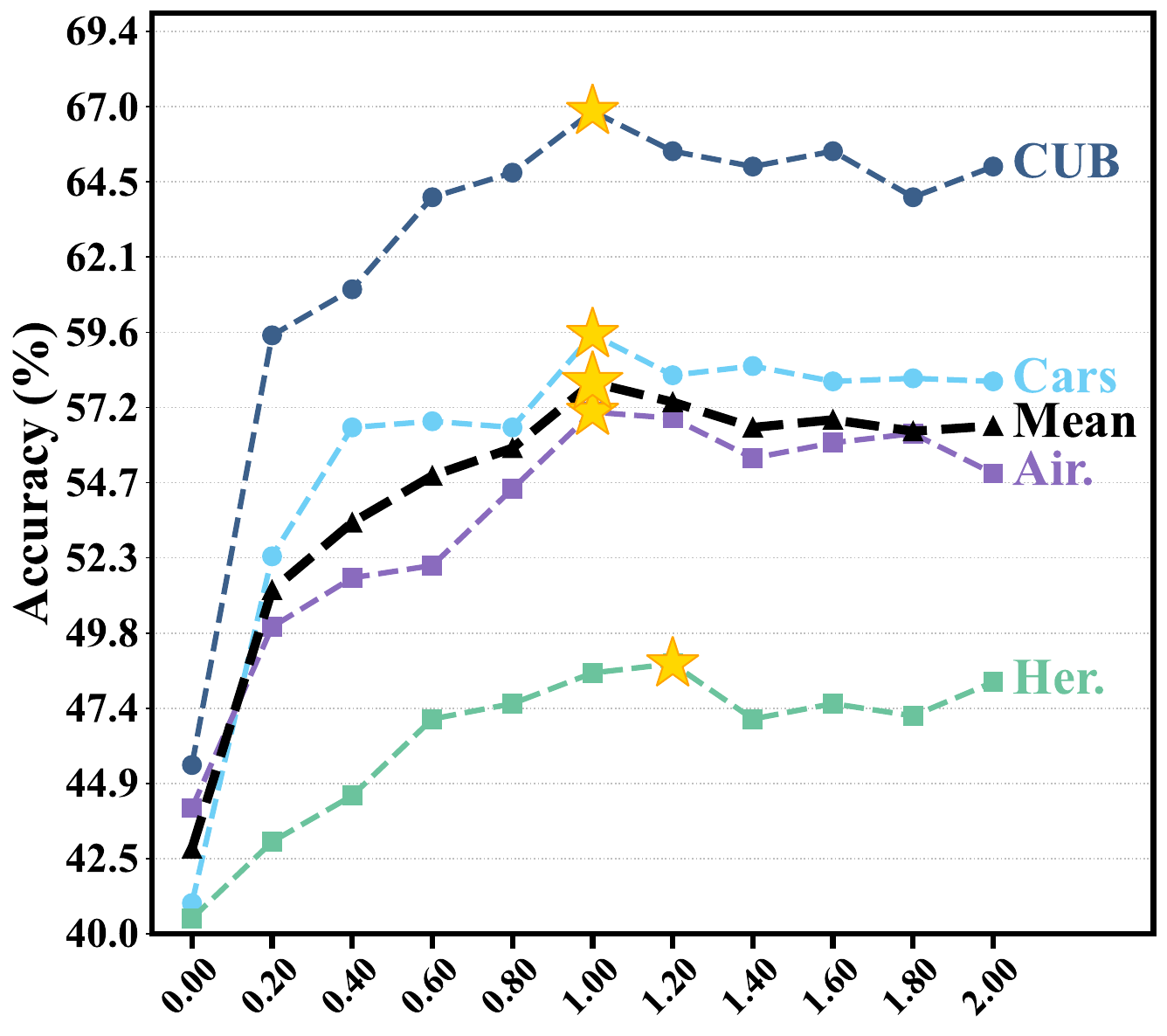}
    \caption{$\lambda_{{inst}}$}
    \label{subfig:lambda_inst}
\end{subfigure}
\hfill
\begin{subfigure}{0.24\textwidth}
    \includegraphics[width=\linewidth]{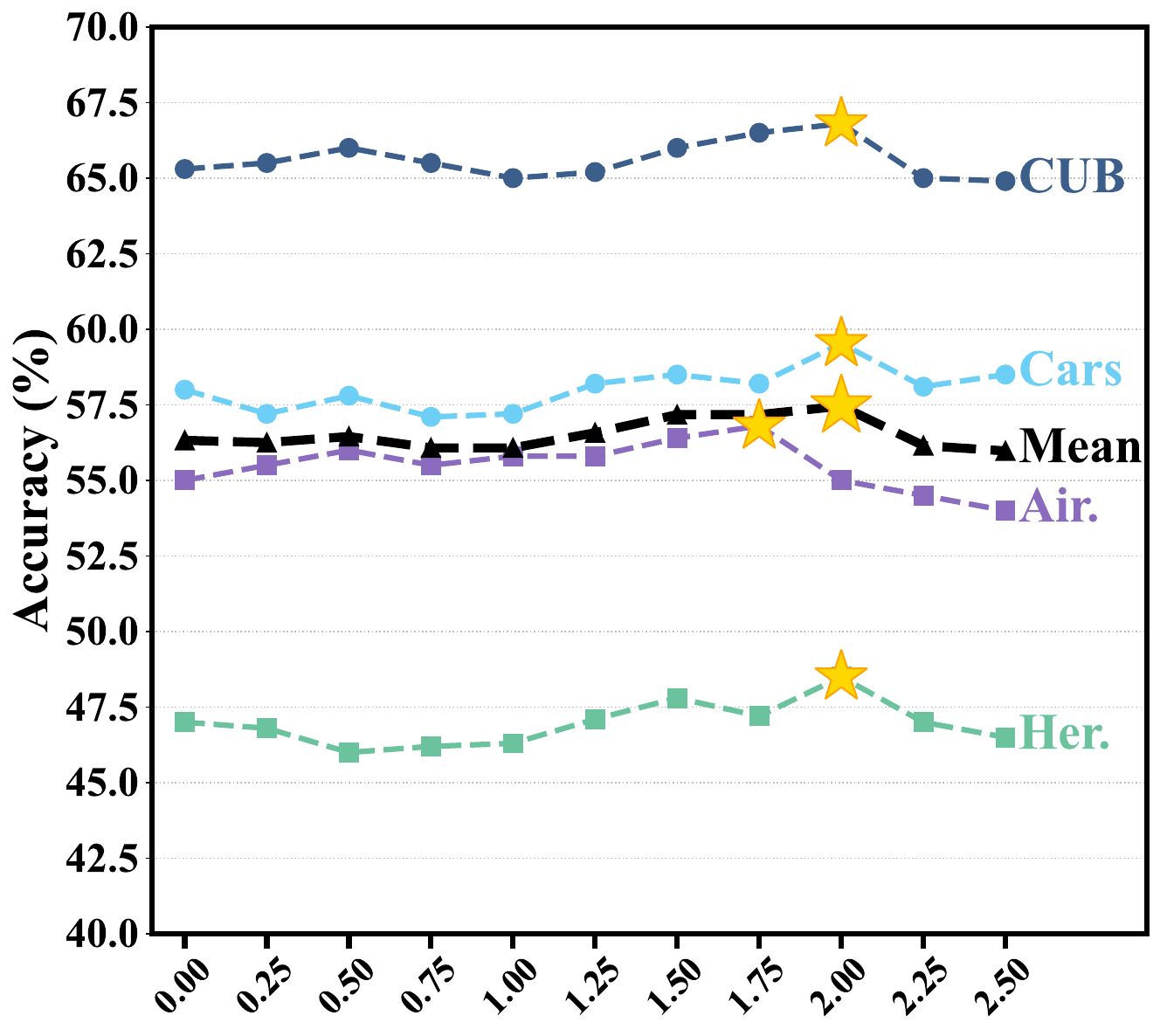}
    \caption{$\lambda_{{ent}}$}
    \label{subfig:lambda_ent}
\end{subfigure}
\caption{Parameter sensitivity analysis for the target hardening parameter ($k$) and loss weights ($\lambda_{\text{cmi}}$, $\lambda_{\text{inst}}$, $\lambda_{\text{ent}}$). The evaluation is conducted on four fine-grained datasets: CUB-200-2011 (CUB), Stanford Cars (Cars), FGVC-Aircraft (Air.), and Herbarium19 (Her.). The dashed black line represents the mean accuracy. Yellow stars indicate the optimal value for each respective curve.}
\label{fig:param_sen}
\end{figure*}

\begin{figure*}
\centering
\begin{subfigure}{0.16\textwidth}
    \includegraphics[width=\linewidth]{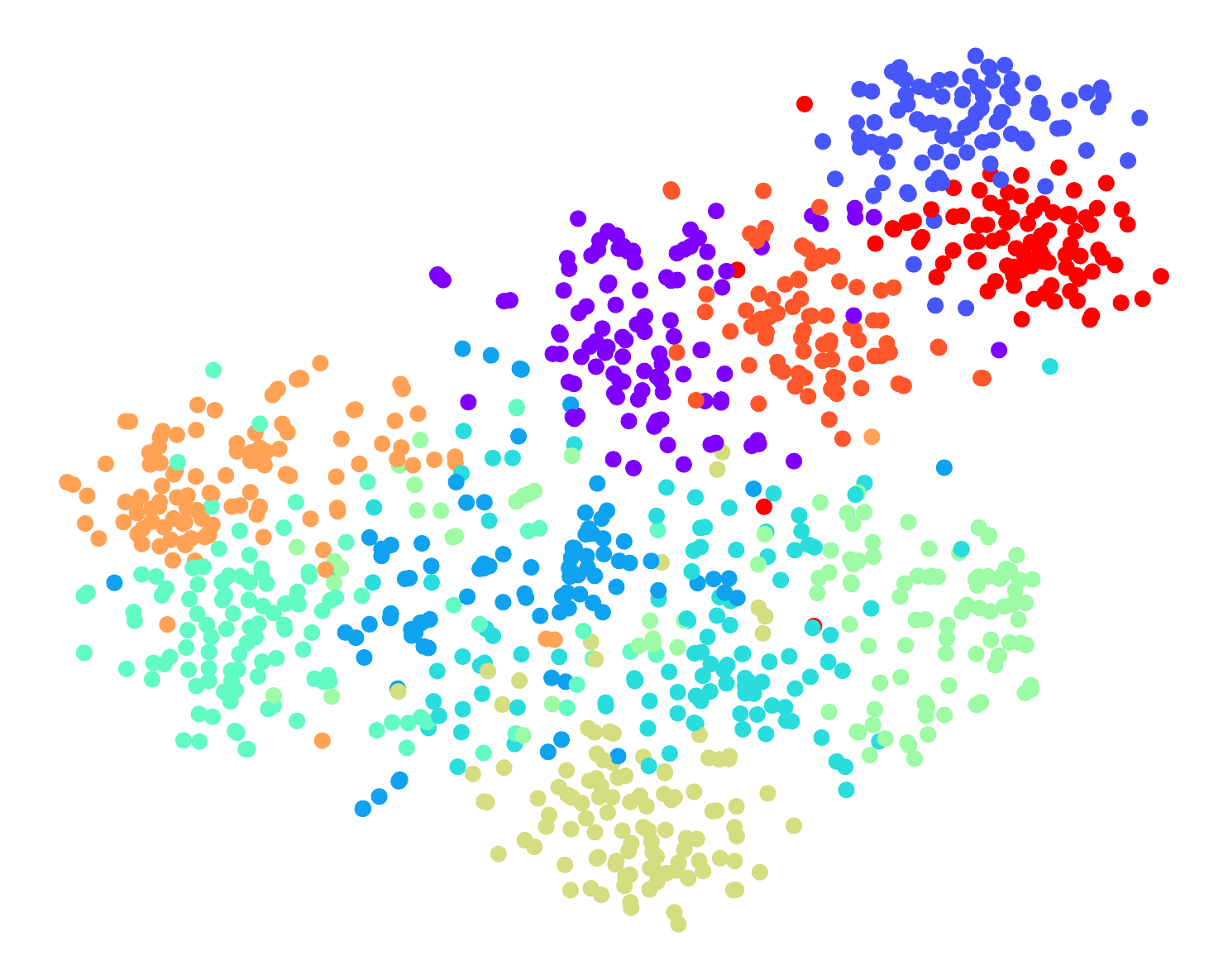}
    \caption{DINO features}
\end{subfigure}
\hfill
\begin{subfigure}{0.16\textwidth}
    \includegraphics[width=\linewidth]{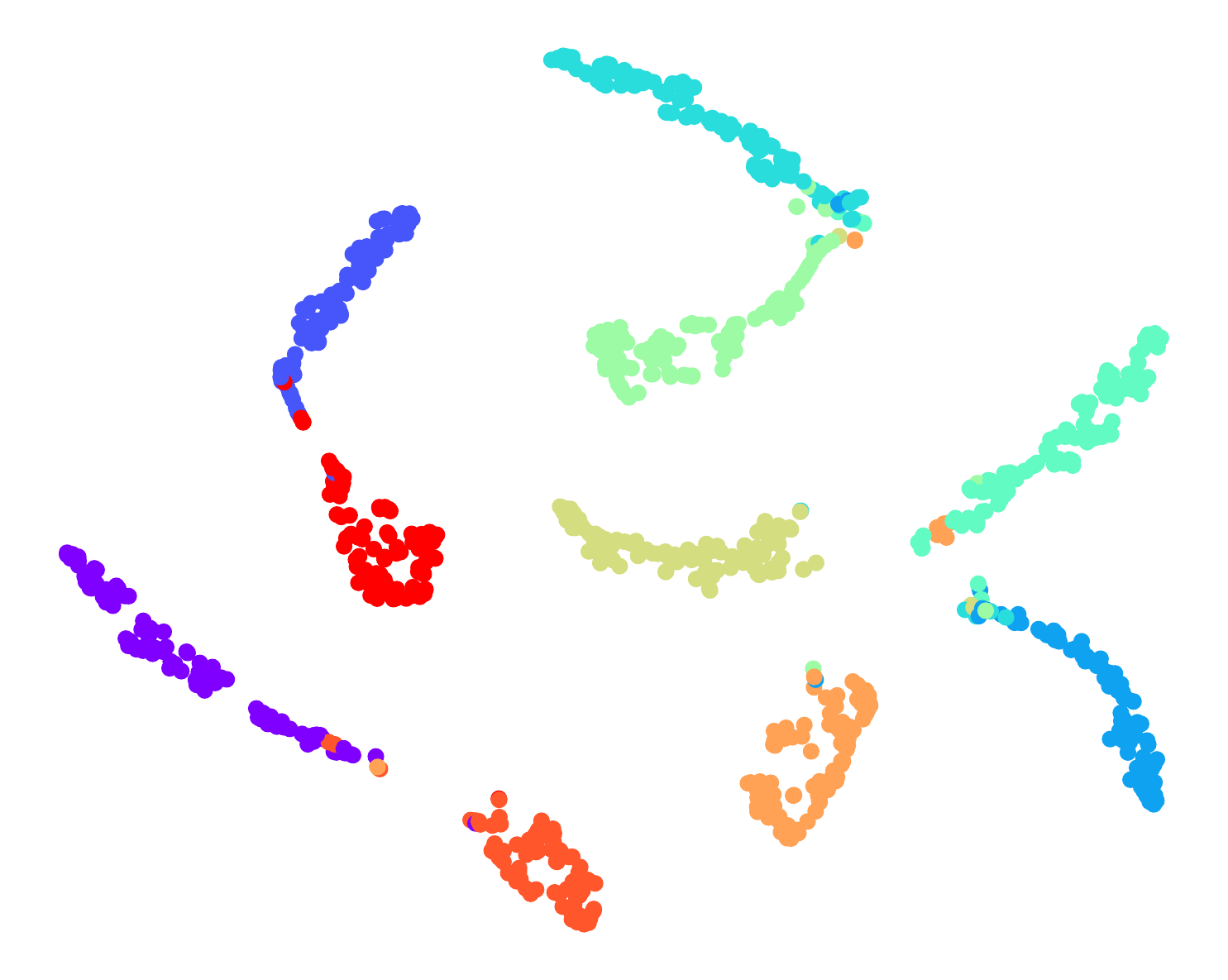}
    \caption{GCD features}
\end{subfigure}
\hfill
\begin{subfigure}{0.16\textwidth}
    \includegraphics[width=\linewidth]{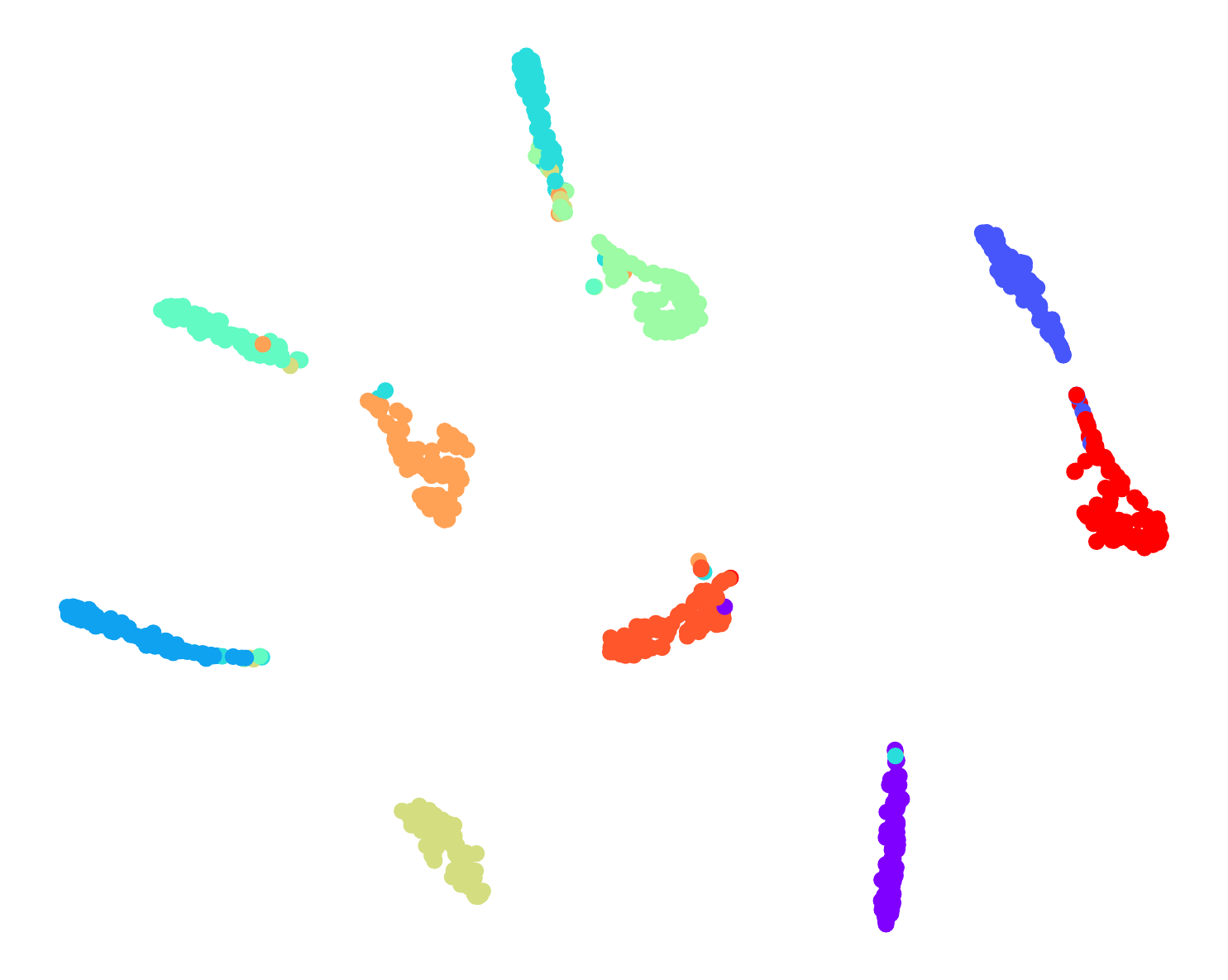}
    \caption{Our features}
\end{subfigure}
\hfill
\rule{0.5pt}{0.1\linewidth}
\hfill
\begin{subfigure}{0.16\textwidth}
    \includegraphics[width=\linewidth]{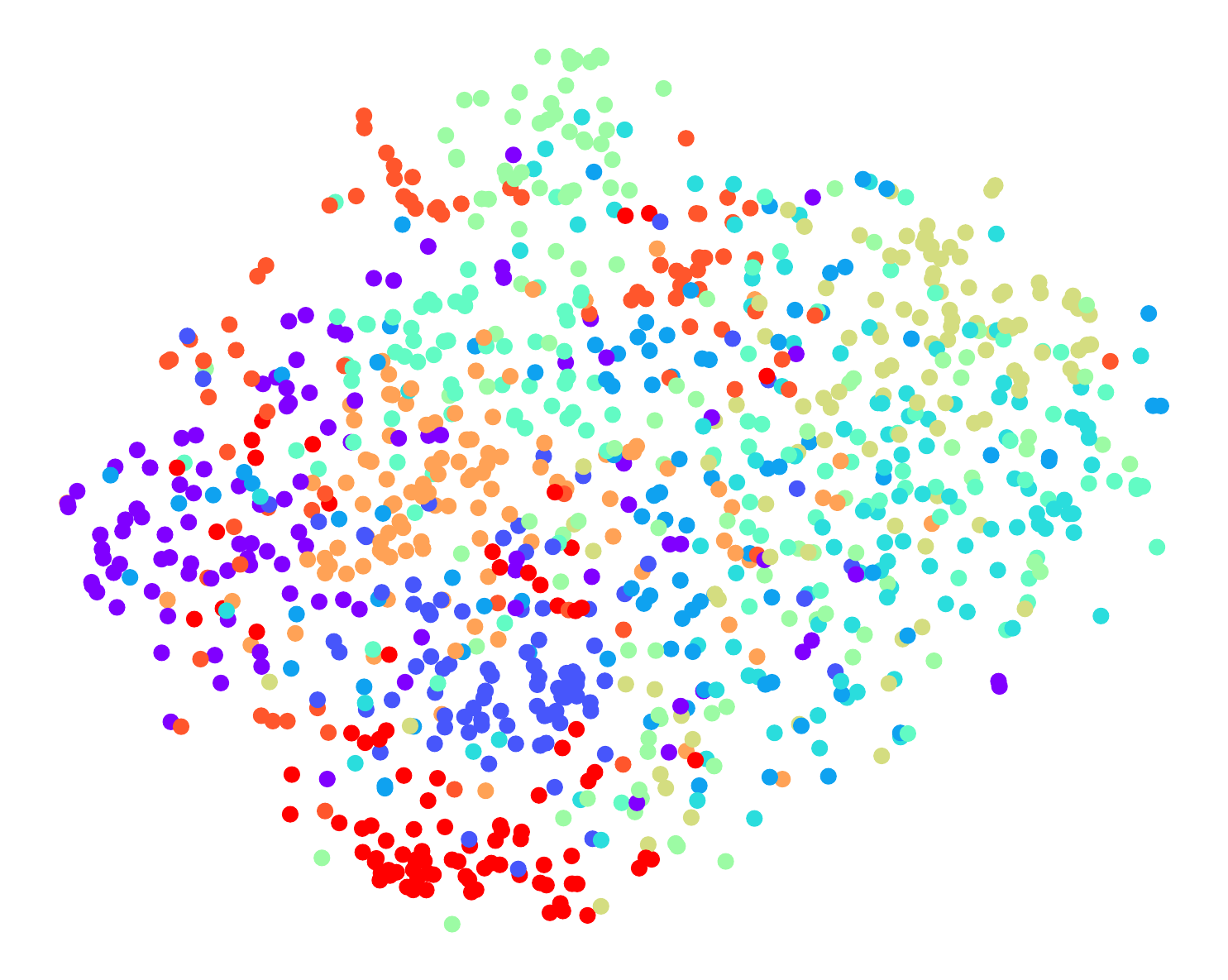}
    \caption{DINO logits}
\end{subfigure}
\hfill
\begin{subfigure}{0.16\textwidth}
    \includegraphics[width=\linewidth]{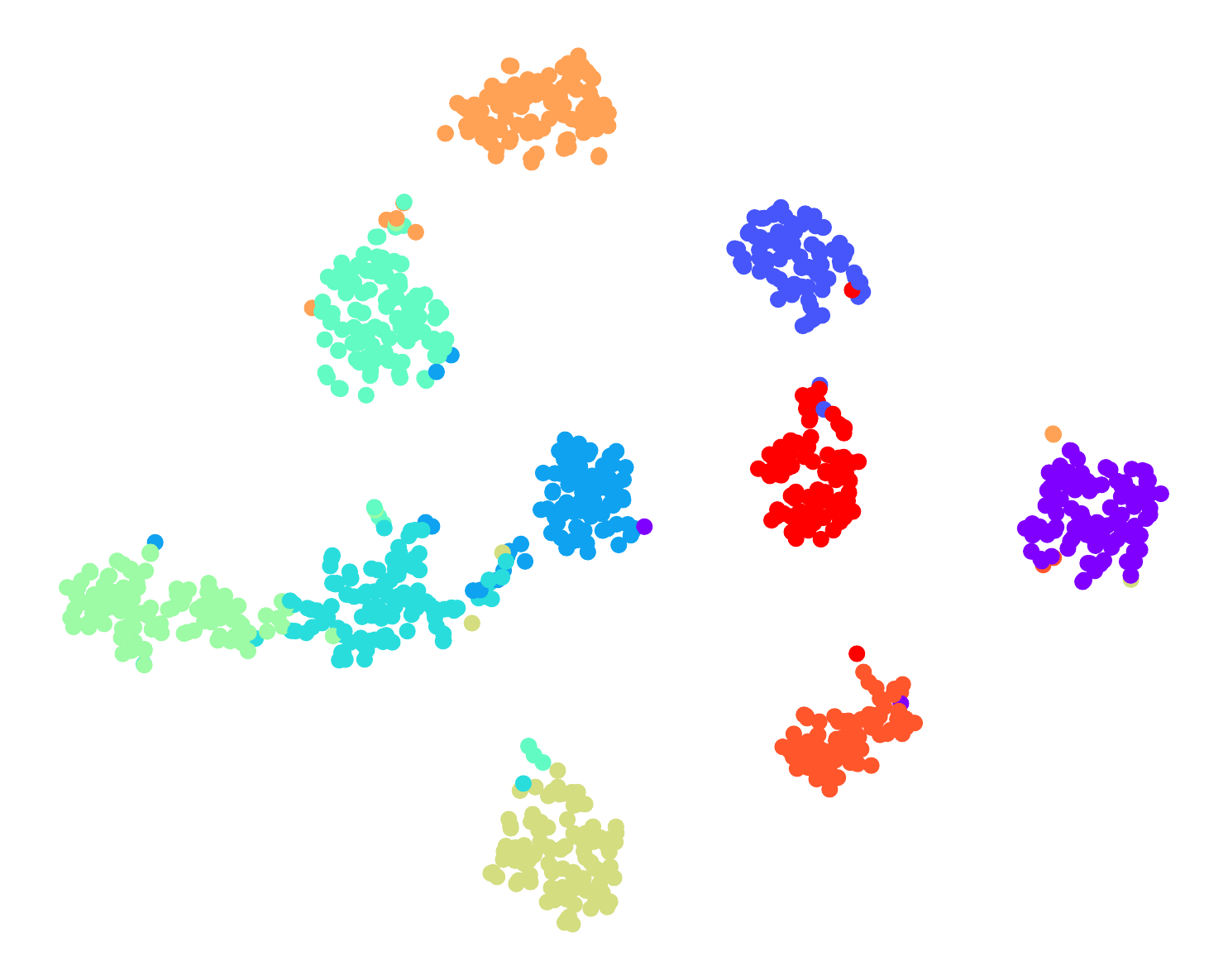}
    \caption{GCD logits}
\end{subfigure}
\hfill
\begin{subfigure}{0.16\textwidth}
    \includegraphics[width=\linewidth]{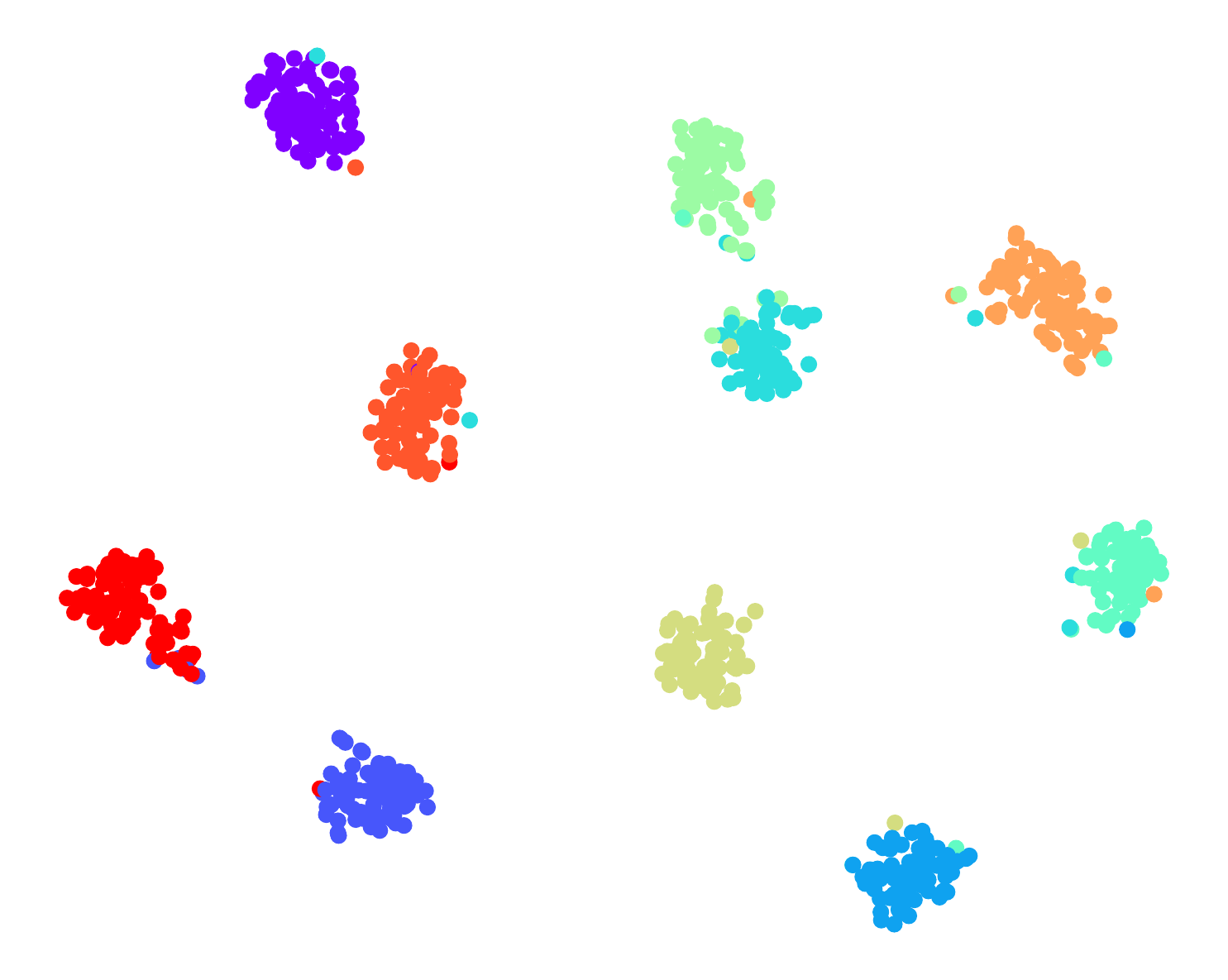}
    \caption{Our logits}
\end{subfigure}
\caption{t-SNE visualizations of feature and logit spaces on the CIFAR-10 test set. }
\label{fig:tsne}
\end{figure*}

\subsubsection{Implementation Details.}

All experiments are trained on a single NVIDIA V100 GPU. We train all models for 200 epochs with a batch size of 128. The hyperparameters are set as follows: $\lambda_{\text{cmi}} = 0.1$, $\lambda_{\text{inst}} = 1$, and $\lambda_{\text{ent}} = 2$. To ensure fairness, we keep $\alpha$ and $\beta$, $\tau_{\text{sep}}$, and $\lambda_{\text{sep}}$ consistent with ProtoGCD \cite{ma2025protogcd}.

\subsection{Comparisions with SOTA methods}

As shown in Table~\ref{tab:main_results_4} and Table \ref{tab:main_results_4_finegrained}, We compare InfoSculpt against SOTA approaches include earlier methods such as GCD, DCCL \cite{pu2023dynamic}, GPC \cite{zhao2023learning}, InfoSieve \cite{rastegar2023learn} and SimGCD, as well as more recent methods including LegoGCD \cite{cao2024solving}, ActiveGCD \cite{ma2024active}, MTMC \cite{tang2025generalized}, Hyp-GCD \cite{liu2025hyperbolic} and ProtoGCD \cite{ma2025protogcd}. 

As demonstrated in Tables~\ref{tab:main_results_4} and~\ref{tab:main_results_4_finegrained}, InfoSculpt achieves state-of-the-art accuracy on both known ("Old") and overall ("All") classes across generic and fine-grained datasets, while maintaining highly competitive performance on novel ones. This holistic improvement is not achieved by narrowly optimizing for novel category discovery, but rather stems from our method's principled approach to sculpting the information space. This is particularly evident on challenging fine-grained datasets. For instance, on Stanford Cars, InfoSculpt's remarkable 78.4\% accuracy on known classes drives its leading 59.5\% overall performance. Similarly, on Herbarium 19, its 48.5\% "All" accuracy represents a significant 2.9\% improvement over the next-best approach. This powerful retention of prior knowledge is a direct validation of our dual CMI objectives, which operate in tandem at different granularities to systematically sculpt a high-quality latent space.

\subsection{Ablation Study}

An ablation study on four fine-grained datasets is performed in Figure~\ref{fig:ablations}. Starting with only supervised cross-entropy (ce), the model performs poorly, especially on novel classes, highlighting the difficulty of GCD. Adding strong representation learning (+rep) via contrastive losses substantially improves performance, while entropy loss (+ent) provides slight stabilization. The most significant gains, however, come from the core CMI-based components, confirming the effectiveness of InfoSculpt.

The introduction of the instance-level CMI loss (+inst) marks the most significant leap in performance for novel category discovery. For instance, on CUB and FGVC-Aircraft, “New" accuracy jumps from under 15\% to over 40\%. This empirically validates the hypothesis that enforcing instance-level invariance is paramount for forming initial, coherent clusters of unknown classes from unlabeled data. Subsequently, incorporating the class-level CMI loss (+cmi) builds directly on this success, further improving both “All" and “Old" accuracy. This demonstrates its effectiveness in refining the feature space by compacting intra-class variance and improving the global cluster structure. Finally, the separation loss (+sep) offers marginal yet consistent gains by explicitly enhancing class separability. Overall, the stepwise analysis confirms that dual-level CMI optimization is the key contributor, effectively improving feature invariance and class separation in GCD.

\subsection{Parameter Sensitivity Analysis}

\label{sec:param_sensitivity}
We conduct a comprehensive sensitivity analysis for four key hyperparameters: the target hardening parameter $k$, the class-level CMI loss weight $\lambda_{\text{cmi}}$, the instance-level CMI loss weight $\lambda_{\text{inst}}$, and the entropy weight $\lambda_{\text{ent}}$. The analysis is performed across four fine-grained datasets with results presented in Figure~\ref{fig:param_sen}.

As shown in Figure~\ref{fig:param_sen}(a), we analyze the effect of $k$, which controls the sharpness of the target distribution. The model's performance is relatively stable and the mean accuracy peaks at $k=10$. This value effectively sharpens the class focus by removing the most distracting negative classes while retaining soft, data-driven knowledge as a regularizer. 

In contrast, The model exhibits greater sensitivity to the CMI loss weights, which are central to our contributions. Figure~\ref{fig:param_sen}(b) shows that for the class-level CMI weight, $\lambda_{\text{cmi}}$, performance steadily improves and peaks at $\lambda_{\text{cmi}}=0.10$. Values larger than this cause a decline in performance, highlighting the need to balance the objective of enhancing intra-class compactness with other learning goals. For the instance-level CMI weight, $\lambda_{\text{inst}}$, Figure~\ref{fig:param_sen}(c) shows a sharp increase in accuracy as the weight moves from 0, with optimal performance achieved at $\lambda_{\text{inst}}=1.0$. This underscores the critical role of promoting invariant representations for fine-grained recognition.

Additionally, Figure~\ref{fig:param_sen}(d) examines the entropy regularization weight, $\lambda_{\text{ent}}$, which prevents model collapse. Similar to $k$, the model is relatively insensitive to this parameter, though a clear peak in mean accuracy occurs at $2.0$, confirming its role in stabilizing learning. Based on these analyses, we adopt the optimal values $k=10$, $\lambda_{\text{cmi}}=0.1$, $\lambda_{\text{inst}}=1.0$, and $\lambda_{\text{ent}}=2.0$ for all main experiments.

\subsection{Visualization and Plug-and-Play Capability}

T-SNE visualizations on the CIFAR-10 test set are used to compare feature and logit spaces across DINO, GCD, and InfoSculpt in Figure~\ref{fig:tsne}. In feature space (Figure~\ref{fig:tsne}a-c), DINO shows overlap, GCD improves separation but lacks compactness, while InfoSculpt achieves both. In logit space (Figure~\ref{fig:tsne}d-f), DINO is disordered, GCD forms loose clusters, and InfoSculpt shows highly compact, well-separated logits with sharper decision boundaries. These results show that dual-level CMI effectively shapes a structured latent space for both known and novel category discovery.

\begin{figure}[H]
    \centering
    \includegraphics[width=0.99\linewidth]{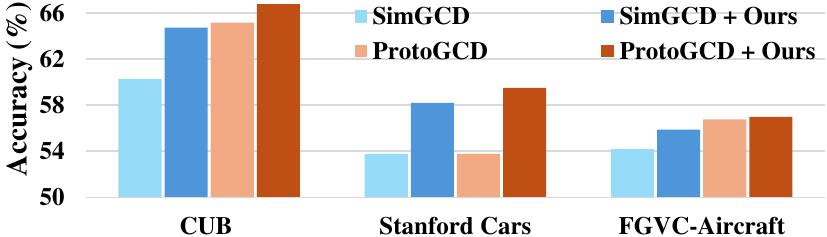}
    \caption{Plug-and-play performance of InfoSculpt.}
    \label{fig:plug}
\end{figure}
To evaluate the generality, the dual CMI framework is integrated into SimGCD and ProtoGCD. As shown in Figure~\ref{fig:plug}, this integration consistently improves performance across three datasets, demonstrating its flexibility and effectiveness as a universal regularization strategy for GCD.

\section{Conclusion}

In this paper, we address a core challenge in GCD: the absence of a principled way to separate category-defining signals from instance-specific noise. We introduce a novel information-theoretic perspective, re-framing the task as a trade-off between representation compressiveness and informativeness. Our proposed \textbf{InfoSculpt} framework instantiates this idea through a dual CMI objective. By combining a Category-Level CMI to structure known classes and an Instance-Level CMI to purify all sample representations, InfoSculpt systematically sculpts a latent space that is robust for both known and novel category discovery. By explicitly controlling information flow, InfoSculpt resolves the core tension in GCD and enables stronger open-world learning.

\bibliographystyle{plain}
\newpage
\bibliography{reference}

\clearpage        
\appendix         
\section*{Appendix} 

\input{sup}

\end{document}

%% file: sup.tex
\crefname{section}{Sec.}{Secs.}
\Crefname{section}{Section}{Sections}
\Crefname{table}{Table}{Tables}
\crefname{table}{Tab.}{Tabs.}

\def\cvprPaperID{***} 
\def\confName{CVPR}
\def\confYear{2026}

\section{Related Works}
\subsection{Generalized Category Discovery}

Proposed by \cite{vaze2022generalized}, GCD addresses the practical challenge of learning from datasets with labeled known classes and unlabeled data containing both known and novel categories. It extends NCD beyond the traditional Semi-Supervised Learning (SSL) \cite{berthelot2019mixmatch} setting to handle mixed-category unlabeled data. Early studies mainly used a non-parametric approach, combining self-supervised backbones with semi-supervised k-means for class partitioning \cite{fei2022xcon}. Despite mitigating overfitting under class imbalance, this approach requires inference-time clustering, making it computationally expensive and unsuitable for instance-wise prediction. To address these issues, research has shifted to parametric classifiers, with SimGCD \cite{wen2023parametric} replacing clustering with a classification head for faster inference and establishing a strong baseline. Follow-up works, such as SPTNet \cite{wang2024sptnet} with spatial prompts and CMS \cite{choi2024contrastive} with hierarchical heads, further enhance feature learning.

\subsection{Conditional Mutual Information}
The IB principle provides a theoretical foundation for learning robust, minimal representations. Recently, CMI has emerged as a powerful and tractable objective within this framework, valued for its dual optimization role. CMI serves dual roles in representation learning. Minimizing CMI acts as a regularizer to enhance semantic consistency by pulling same-class samples closer, improving classification \cite{yang2025conditional} and addressing class imbalance \cite{hamidi2024fed}. Conversely, maximizing CMI facilitates alignment across models or distributions, benefiting knowledge distillation \cite{yang2024conditional} and student-teacher regularization \cite{yang2024markov}.

These prior works demonstrate the effectiveness of CMI in sculpting the feature space for specific tasks. Building on this insight, InfoSculpt is the first to systematically introduce and adapt the CMI framework to tackle the multifaceted challenges of GCD, enhancing generalization and feature separability in open-world scenarios.

\section{Detailed Theoretical Support}

This section aims to provide a more detailed exposition of the theoretical foundations, include: \begin{itemize}
    \item[\ref{IB2CMI}] A complete derivation from the Information Bottleneck (IB) principle to our tractable Conditional Mutual Information (CMI) objective;
    \item[\ref{markov}] An in-depth discussion of the core assumption, the Markov chain $Y \rightarrow X \rightarrow \hat{Y}$;
    \item[\ref{loss}] A particular emphasis on the refined target distribution $\boldsymbol{\hat{q}}^{y_i}$, and a further elaboration on the design principles of $\mathcal{L}_{\text{inst}}$.
\end{itemize}

\subsection{Detailed Derivation from IB to CMI}
\label{IB2CMI}

The main paper posits that minimizing the CMI, $I(X; \hat{Y}|Y)$, is equivalent to a specialized and tractable formulation of the IB principle. Here, we provide a more formal justification for this claim.

The IB principle seeks to learn a compressed representation $Z$ of an input $X$ by minimizing the Lagrangian:
\begin{equation}
    \mathcal{L}_{\text{IB}} = I(X; Z) - \beta I(Y; Z)
\end{equation}
where minimizing $I(X; Z)$ enforces \textbf{compression} (discarding information from $X$ that is irrelevant to the label $Y$), and maximizing $I(Y; Z)$ ensures \textbf{sufficiency} (retaining information that is predictive of $Y$).

Inspired by \cite{zhuang2025stealthy}, we designate the model's output logits, represented by the random variable $\hat{Y}$, as the compressed representation, i.e., $Z = \hat{Y}$. Our objective is to minimize the CMI, which is defined as:
\begin{equation}
    I(X; \hat{Y}|Y) = H(\hat{Y}|Y) - H(\hat{Y}|X, Y)
\end{equation}
By invoking the Markov chain $Y \rightarrow X \rightarrow \hat{Y}$ (justified in Section \ref{markov}), we have $H(\hat{Y}|X, Y) = H(\hat{Y}|X)$. The CMI expression thus simplifies to:
\begin{equation}
    I(X; \hat{Y}|Y) = H(\hat{Y}|Y) - H(\hat{Y}|X)
\end{equation}
With the definition of mutual information, $I(A; B) = H(A) - H(A|B)$, we can rewrite the terms above as:
\begin{align}
    I(X; \hat{Y}\mid Y) 
    &= \left(H(\hat{Y}) - I(Y; \hat{Y})\right) \nonumber \\
    &\quad - \left(H(\hat{Y}) - I(X; \hat{Y})\right) \\
    &= I(X; \hat{Y}) - I(Y; \hat{Y})\\
    &= I(X; Z) - I(Y; Z)\\
    &= \mathcal{L}_{\text{IB}}\big|_{\beta = 1}
\end{align}

Therefore, \textbf{minimizing the CMI $I(X; \hat{Y}|Y)$ is equivalent to simultaneously minimizing $I(X; \hat{Y})$ and maximizing $I(Y; \hat{Y})$}.

This objective aligns perfectly with the two goals of the IB principle:
\begin{enumerate}
    \item \textbf{Minimizing $I(X; \hat{Y})$} corresponds to the \textbf{compression} term. It encourages the model's prediction $\hat{Y}$ to be as independent as possible from the specific input instance $X$. This forces the model to learn class-level, generalizable features rather than memorizing instance-specific artifacts.
    \item \textbf{Maximizing $I(Y; \hat{Y})$} corresponds to the \textbf{sufficiency} term. It requires the model's prediction $\hat{Y}$ to be maximally informative about the true label $Y$, thus ensuring predictive accuracy.
\end{enumerate}
Compared to the classic IB formulation, the objective $I(X; \hat{Y}) - I(Y; \hat{Y})$ can be viewed as a special case with $\beta=1$. It operates directly in the logit space, circumventing the notoriously difficult task of estimating mutual information for high-dimensional, continuous representations, which makes it a more stable and practical objective for deep neural networks.

\subsection{Discussion on Markov Chain Assumption}
\label{markov}

Our derivation of the practical CMI estimator hinges on the assumption of the Markov chain $Y \rightarrow X \rightarrow \hat{Y}$. We discuss its justification and potential limitations here.

\paragraph{Justification.} This Markov chain posits that, given the input $X$, the model's prediction $\hat{Y}$ is conditionally independent of the true label $Y$. This perfectly describes the standard inference process of a supervised classifier: the model receives an input $X$ and produces a prediction $\hat{Y}$ based solely on its parameters and the information contained within $X$. During this process, the model has no direct access to the ground-truth label $Y$. The flow of information is thus unidirectional: $Y$ influences the generation of $X$ (e.g., the label "cat" causes an image $X$ to contain a cat), which in turn determines the model's prediction $\hat{Y}$ \cite{tishby2000information}.

\paragraph{Potential Limitations.} In certain scenarios, this assumption could be weakly violated. For instance, with Batch Normalization \cite{ioffe2015batch}, the statistics for a single sample are influenced by other samples in the same mini-batch. If the label distribution within a batch is skewed, a faint trace of information about $Y$ could "leak" to $\hat{Y}$ through the batch statistics \cite{ioffe2017batch}. However, this effect is generally considered negligible noise.

\subsection{Further Elaboration on the Loss Design}
\label{loss}

\subsubsection{Motivation for Refining Target Distribution}

The main text describes the construction of a refined target $\boldsymbol{\hat{q}}^{y_i}$ from the empirical class centroid $\boldsymbol{q}^{y_i}$. The rationale behind this design is threefold:

\textbf{Strengthening the Supervisory Signal:} The raw centroid $\boldsymbol{q}^{y_i}$ is an average over a finite number of predictions and may be noisy or lack "sharpness" (i.e., its confidence in the ground-truth class may be well below 1). By manually setting the confidence for the true class $y_i$ to 1, we provide the model with an unambiguous and powerful supervisory signal, guiding its convergence more effectively.
    
\textbf{Hard-Negative Suppression:} In a multi-class problem, a sample can share visual similarities with several incorrect classes. The non-ground-truth classes with the highest confidence in the centroid $\boldsymbol{q}^{y_i}$ often represent the "hardest" negatives—those most easily confused with the true class. By zeroing out the top-$k$ of these confidences, we explicitly instruct the model to increase the separation in its decision boundary against these specific challenging classes.
    
\textbf{Preserving Soft Knowledge and Preventing Overfitting:} We only suppress the top-$k$ hardest negatives, preserving the remaining low-confidence values. This faint signal can be interpreted as "soft knowledge" that reflects latent inter-class similarities in the data (e.g., "trucks" are more similar to "cars" than to "birds"). Retaining this information acts as a form of regularization, akin to Label Smoothing, which prevents the model from making over-confident negative predictions and improves its generalization capability.

\subsubsection{Information-Theoretic View of Instance-Level Consistency Loss}

The instance-level consistency loss, $\mathcal{L}_{\text{inst}}$, is designed to leverage the entire dataset $\mathcal{D}$ by enforcing that the model's predictions are robust to data augmentations. We can frame this objective from an information-theoretic perspective, where the goal is to learn a prediction that is \textbf{invariant} to the instance-specific details of an augmented view while remaining \textbf{informative} about the core identity of the underlying sample.

Let us denote an underlying data sample as $X_{inst}$ and a specific augmented view of it as $X_{aug}$. The goal of learning an invariant prediction can be formalized as minimizing the conditional mutual information between the augmented view and the prediction, given the instance identity $\min I(X_{aug}; \hat{Y} | X_{inst})$.

Intuitively, this objective states that once we know the identity of the sample ($X_{inst}$), the specific augmentation used to generate the view ($X_{aug}$) should provide no additional information about the model's prediction ($\hat{Y}$). Minimizing this CMI encourages the model to discard features that are specific to the augmentation and rely only on features that represent the core essence of the instance.

Following the derivation in the main text, 
\begin{equation} 
    I(X_{aug}; \hat{Y} | X_{inst}) = \frac{1}{n} \sum_{i=1}^{n} \mathrm{KL}\left( \boldsymbol{p}_i^{aug} \,\|\, \boldsymbol{p}_i^{inst} \right).
\end{equation}

To create a tractable estimator, we employ a strategy common in self-supervised learning. We use a different augmented view of the same instance as a \textbf{proxy} for the ideal, invariant target. Let $\boldsymbol{x}_i$ and $\boldsymbol{x}_i'$ be two distinct augmentations of the same underlying sample, with corresponding softmax predictions $\boldsymbol{p}_i$ and $\boldsymbol{p}_i'$. As stated in the main text, the stop-gradient prediction from one view, $\boldsymbol{q}_i'$, serves as the target for the other view's prediction, $\boldsymbol{p}_i$. Symmetrically, The stop-gradient prediction $\boldsymbol{q}_i$ from the first view also serves as the target for the second view's prediction, $\boldsymbol{p}_i'$. This results in the final symmetric loss, which matches Equation (17) in the main text:
\begin{equation}
    \mathcal{L}_{\text{inst}} = \frac{1}{2|\mathcal{B}|} \sum_{i \in \mathcal{B}} \left( \mathrm{KL}(\boldsymbol{q}_i' \,\|\, \boldsymbol{p}_i) + \mathrm{KL}(\boldsymbol{q}_i \,\|\, \boldsymbol{p}_i') \right).
\end{equation}

In this way, $\mathcal{L}_{\text{inst}}$ directly operationalizes the CMI minimization principle at the instance level. By forcing the predictions from two different views to be close, it compels the model to "compress away" the information unique to each augmentation while retaining the shared information that defines the instance's identity.

Notably, the $\mathcal{L}_{\text{inst}}$ eschews the refining procedure used in category-level loss $\mathcal{L}_{\text{cmi}}$. The objective is not to enforce a hard pseudo-label, but rather to align the complete predictive distributions from two views. By avoiding refining, we preserve a richer supervisory signal, promoting a closer match between the continuous logits.

\section{Limitation and Future Work}

While InfoSculpt demonstrates strong performance, its validation is currently centered on image classification. A key avenue for future work is to extend this information-theoretic framework to more complex tasks, such as open-world object detection and semantic segmentation, where the challenge of disentangling salient signals from noise is more pronounced. 